\documentclass[letterpaper, 10 pt, conference]{ieeeconf}  % Comment this line out if you need a4paper

\IEEEoverridecommandlockouts                              % This command is only needed if 
\usepackage{amsmath}
\usepackage{amssymb}
\usepackage{graphicx}
\usepackage{amsfonts}
\usepackage{subfigure}
\usepackage{cite}
\usepackage{verbatim}

\usepackage{amsthm}
\usepackage{array}
\usepackage{caption}
\usepackage{color}
\let\labelindent\relax
\usepackage{enumitem}

\title{\LARGE \bf
Limits of Probabilistic Safety Guarantees when Considering Human Uncertainty
}

\author{Richard Cheng$^{1}$, Richard M. Murray$^{1}$, and Joel W. Burdick$^{1}$% <-this % stops a space
\thanks{$^{1}$All authors are with the California Institute of Technology, Pasadena, CA, USA
        {\tt\small rcheng@caltech.edu, murray@cds.caltech.edu, jwb@robotics.caltech.edu}}%
}

\begin{document}

\maketitle
\thispagestyle{empty}
\pagestyle{empty}

%%%%%%%%%%%%%%%%%%%%%%%%%%%%%%%%%%%%%%%%%%%%%%%%%%%%%%%%%%%%%%%%%%%%%%%%%%%%%%%%
\begin{abstract}
When autonomous robots interact with humans, such as during autonomous driving, explicit safety guarantees are crucial in order to avoid potentially life-threatening accidents. Many data-driven methods have explored learning probabilistic bounds over human agents' trajectories (i.e. confidence tubes that contain trajectories with probability $\delta$), which can then be used to guarantee safety with probability $1-\delta$. However, almost all existing works consider $\delta \geq 0.001$. The purpose of this paper is to argue that \textit{(1)} in safety-critical applications, it is necessary to provide safety guarantees with $\delta < 10^{-8}$, and \textit{(2)} current learning-based methods are ill-equipped to compute \textit{accurate} confidence bounds at such low $\delta$. Using human driving data (from the \textit{highD} dataset), as well as synthetically generated data, we show that current uncertainty models use inaccurate distributional assumptions to describe human behavior and/or require infeasible amounts of data to accurately learn confidence bounds for $\delta \leq 10^{-8}$. These two issues result in unreliable confidence bounds, which can have dangerous implications if deployed on safety-critical systems. % We conclude by suggesting that potential solutions incorporate prior knowledge of human interaction behavior, encoded through assume-guarantee contracts. 
\end{abstract}

%%%%%%%%%%%%%%%%%%%%%%%%%%%%%%%%%%%%%%%%%%%%%%%%%%%%%%%%%%%%%%%%%%%%%%%%%%%%%%%%
\section{INTRODUCTION}

Autonomous robots will be increasingly deployed in unstructured human environments (e.g. roads and malls) where they must safely carry out tasks in the presence of other moving human agents. The cost of failure is high in these environments, as safety violations can be life-threatening. At present, safety is often enforced by learning an uncertainty distribution or confidence bounds over the future trajectory of other agents, and designing a controller that is robust to such uncertainty \cite{Fisac2018}. Based on these learned trajectory distributions, probabilistic safety guarantees can be provided at a specified safety threshold $\delta$ over a given planning horizon (e.g. by enforcing chance constraints such that $\mathbb{P}(\text{collision}) \leq \delta$) \cite{Aoude2013,FridovichKeil2020,Nakka2020,Fan2020}. However, for such guarantees to hold, it is critical that we \textit{accurately predict} the uncertainty over other agents' future trajectories with high probability $1-\delta$.

% 3,240 billion miles, 5,215 billion km, 6,734,000 crashes, 1,928,000 injuries

% \note{The italicized quote below is poorly formatted}

Current works that aim to provide probabilistic safety guarantees for autonomous navigation in uncertain, human environments consider safety thresholds in the range $\delta \geq 0.001$. While such guarantees are important, safety critical applications require $\delta$ that are orders of magnitude lower \cite{Shalev-Shwartz2017}.
\vspace{-0.9em}
\begin{quotation}
\noindent
\textit{Suppose a robot/car is guaranteed safe with probability $1-\delta$ across every $10s$ planning horizon. Given $\delta \approx 0.001$, we could expect a safety violation every 3 hrs. For reference, based on NHTSA data \cite{nhtsa}, human drivers have an effective safety threshold $\delta < 10^{-7}$}. 
\end{quotation}
\vspace{-0.2em}

It is clear then that for safety-critical robotic applications, we must strive for extremely low safety thresholds, on the order $\delta \leq 10^{-8}$. However, this paper argues that current learning-based approaches that model human trajectory uncertainty \textit{(a)} rely on highly inaccurate distribution assumptions, invalidating resulting safety guarantees, and/or \textit{(b)} can not adequately extend to safety-critical situations. To illustrate this, we applied different uncertainty models (see Table 1) to data of human driving from the \texttt{highD} dataset \cite{highDdataset}. We found that even under extremely generous assumptions, learned models are highly inaccurate in capturing human behavior at low $\delta$, often mispredicting the probability of rare events by several orders of magnitude. Furthermore, we show that increasing dataset sizes will not sufficiently improve accuracy of learned uncertainty models. % In addition to the real-world datasets, we support our conclusions with simulations on synthetic data, though these results are pushed to the Appendix.

% (1) our results are not focused on safe/robust control algorithms, but rather the learned (or assumed) uncertainties that they must leverage. (2) We are focused on aleatoric uncertainty, rather than epistemic uncertainty. In other words, we highlight the inability of any class of model uncertainty (regardless of learning algorithm and generalization) to accurately capture distributions at low $\delta$.  % In other words, we are \textit{not} considering the ability of algorithms to learn generalizable uncertainty distributions, but rather the inability of any class of model uncertainty (regardless of learning algorithm and generalizability) to 

Our results highlight potential danger in utilizing learned models of human uncertainty in safety-critical applications. Fundamental limitations prevent us from accurately learning the probability of rare trajectories with finite data, and using inaccurate confidence bounds can result in unexpected collisions. While this paper focuses on illustrating a crucial problem (rather than providing a solution), we conclude by discussing alternative approaches that can address these limitations by combining \textit{(a)} learned patterns of behavior and \textit{(b)} prior knowledge encoding human interaction rules.% an alternate approach utilizing assume-guarantee contracts \cite{phan2019}; we believe these tools can address the limitations pointed out in this work by rigorously combining \textit{(a)} learned patterns of behavior and \textit{(b)} prior knowledge encoding human interaction rules.

Before proceeding, we emphasize three critical points regarding our results:
\begin{itemize}
    \item We focus on in-distribution error, rather than out-of-distribution error. I.e., we highlight the fundamental inability of uncertainty models to accurately capture distributions at very low $\delta$, regardless of generalization.
    \item We focus not on robust control algorithms, but rather on the learned uncertainties that such algorithms leverage.
    \item We distinguish \textit{motion predictors} from \textit{uncertainty models}. While recent performance of motion predictors has drastically improved \cite{Gupta2018}, they all leverage an underlying uncertainty model (see Table 1) to capture the probability of uncommon events. E.g. most neural network motion predictors output a Gaussian uncertain prediction. This paper focuses on errors associated with uncertainty models (which propagate to the motion predictors).
\end{itemize}

% Section II reviews the literature on human uncertainty modeling for safe motion planning. In Section III, we highlight issues with these models for safety-critical applications. In Section IV, we conclude by proposing handling human uncertainty through the lens of assume-guarantee contracts. 

%\begin{table}[!h]
\begin{table*}[t]
    \centering
    \normalsize
    \begin{tabular}{| m{5.0cm} | m{2.8cm}| m{3.3cm} |}
        \hline
        Uncertainty Model Class & Example Works & Min. Safety Threshold \\ % [0.5ex] 
        \hline\hline
        Gaussian Process &  \cite{Fisac2019,Aoude2013,Hakobyan2020,Cheng2020} & $\delta \geq 0.001$  \\
        \hline
        Gaussian Uncertainty w/ Dynamics & \cite{Xu2014,Sadigh2016a,forghani2016} & $\delta \geq 0.001$ \\
        \hline
        Bayesian NN & \cite{Michelmore2019,Fan2020} & $\delta \geq 0.05$ \\
        \hline
        Noisy Rational Model & \cite{FridovichKeil2020} & $\delta \geq 0.01$ \\
        \hline
        Hidden Markov Model & \cite{sadigh2014,Liu2015} & $\delta \geq 0.01$ \\
        \hline
        Quantile Regression & \cite{Fan2020} & $\delta \geq 0.05$ \\
        \hline
        Scenario Optimization & \cite{Cesari2017,Chen2020,Sartipizadeh2020} & $\delta \geq 0.01$ \\
        \hline
        Generative Models (e.g. GANs) & \cite{Gupta2018,Sadeghian2019,Salzmann2020} & N/A \\
        \hline
    \end{tabular}
    \caption{Different model classes for capturing human trajectory uncertainty, used in safe planning algorithms to guarantee safety with probability $1-\delta$. The right column shows the lowest safety threshold, $\delta$, we could find used in the literature (in simulation or hardware experiments) for each model class. There is no entry for generative models, as these models have not yet been utilized to provide \textit{explicit} safety guarantees during planning, though there is surely a trend in this direction.}
    \label{table:models}
    % \vskip -0.25 true in
\end{table*}
%\end{table}

\section{RELATED WORK}

% Some works model humans using dynamics models with learned parameters, and assume some uncertainty is captured as a bounded disturbance on that model \cite{Liu2017,Leung2020}. While such approaches may be able to guarantee safety under the assumed disturbances, they rely on overly strong assumptions on the uncertainty. Therefore, most recent approaches for guaranteed safe navigation around humans approximate uncertainty in human trajectories as a random process (i.e. deviations from a nominal trajectory are drawn i.i.d. from some fixed or varying distribution). These uncertainty models help capture noise and the effects of unobserved variables (e.g. intention), and enable probabilistic guarantees on safety in uncertain, dynamic environments. Most of these models fall into one (or more) of the following categories:

Most recent approaches for guaranteed safe navigation in proximity to humans or their cars approximate uncertainty in human trajectories as a random process (i.e. deviations from a nominal trajectory are drawn i.i.d. from a learned distribution). These uncertainty models capture noise and the effects of latent variables (e.g. intention), and enable probabilistic safety guarantees in uncertain, dynamic environments. Most models fall into one or more of the following categories:

\vspace{-0.13em}
\begin{itemize}[leftmargin=*]
    \item \textbf{Gaussian Process (GP):} These approaches model other agents' trajectories as Gaussian processes, which treat trajectory uncertainty as a multivariate Gaussian \cite{Ellis2009,Aoude2013,Hakobyan2020,Cheng2020}. There are several extensions, such as the IGP model \cite{Trautman2015} (which accounts for interaction between multiple agents), or others \cite{ferguson2015,liu2019}. However, they all treat uncertainty as a multivariate Gaussian.
    \item \textbf{Gaussian Noise with Dynamics Model:} These approaches use a dynamics model with additive Gaussian noise; noise can also be added in state observations. This induces a Gaussian distribution over other agents' future trajectory (or a situation where we can do moment-matching) \cite{gray2013,forghani2016}.
    \item \textbf{Quantile Regression:} This approach computes quantile bounds over the trajectories of other agents at a given confidence level, $\delta$. This approach benefits from not assuming an uncertainty distribution over trajectories \cite{Tagasovska2018,Fan2020}. 
    \item \textbf{Scenario Optimization:} This approach computes a predicted set over other agents' actions based on samples of previously observed scenarios \cite{Campi2018}. It is distribution-free (i.e. does not assume a parametric uncertainty distribution) \cite{carvalho2015,Cesari2017,Chen2020,Sartipizadeh2020}. \cite{driggs-campbell2017,driggs-campbell2018} do not use scenario optimization, but their work based on computing minimum support sets follows a similar flavor.
    % \item \textbf{Scenario Optimization:} This approach computes a predicted set over other agents' actions and the probability that their action will fall within that set \cite{Campi2018}. This is a distribution-free approach (i.e. does not compute the uncertainty distribution) \cite{carvalho2015,Cesari2017,Chen2020,Sartipizadeh2020}. \cite{driggs-campbell2017,driggs-campbell2018} do not use scenario optimization, but their method based on computing minimum support sets follows a similar flavor.
    \item \textbf{Noisy (i.e. Boltzmann) Rational Model:} This model treats the human as a rational actor who takes ``noisily optimal'' actions according to a distribution in the exponential family, shown in Eq. \eqref{eq:noisy_rational}. The uncertainty in the action is captured by this distribution, which relies on an accurate model of the human's value function \cite{Li2016,Sadigh2016,Fisac2018,FridovichKeil2020,Kwon2020}.
    \item \textbf{Generative Models (CVAE, GAN):} These models generally learn an implicit distribution over trajectories. Rather than give an explicit distribution, they generate random trajectories that attempt to model the true distribution \cite{Gupta2018,Sadeghian2019}. However, other works have also utilized the CVAE framework to produce \textit{explicit} parameterized distributions using a discrete latent space \cite{Salzmann2020}. % We discuss these models due to their prevalence in trajectory prediction, though to our knowledge, they have not been utilized to provide explicit safety guarantees under uncertainty.
    \item \textbf{Hidden Markov Model (HMM) / Markov Chain:} These models capture uncertainty over \textit{discrete} sets of states/intentions (e.g. goal positions) -- as opposed to capturing uncertainty over trajectories. Thus, the objective is to infer the other agents' unobserved state/intention (from a discrete set) with very high certainty, $1-\delta$ \cite{Kelley2008,McGhan2015,Bandyopadhyay2013,Lam2014,Tran2015,sadigh2014,Liu2015}. % These approaches typically assume approximately known dynamics of the agent but an unobserved state, which must be inferred from observations \cite{Kelley2008,McGhan2015,Bandyopadhyay2013,Lam2014,Tran2015,sadigh2014,Liu2015a,Liu2015}. % This model is used to capture uncertainty over intentions rather than trajectories, but can be coupled with the other models above to incorporate trajectory uncertainty.
    \item \textbf{Uncertainty Quantifying (UQ) Neural Networks:} These approaches do not constitute a separate class of uncertainty models, but refer to methods that train a neural network to capture the distribution over other agents' trajectories \cite{Gindele2010,Kahn2017,Fan2019,Michelmore2019}. We list them separately due to their popularity. Most often these networks output a Gaussian distribution or mixture of Gaussians (e.g. Bayesian neural networks \cite{Blundell2015}, deep ensembles \cite{Lakshminarayanan2016}, Monte-Carlo dropout \cite{Gal2016}). These models can also quantify uncertainty over discrete states (i.e. infer the hidden state in HMMs) \cite{Hu2018,Ding2019}.
\end{itemize}
Once a predicted trajectory and its uncertainty is learned, many mechanisms exist to guarantee safety (e.g. incorporating uncertainty into chance constraints). In this work, we do not focus on these mechanisms (i.e. robust control algorithms) for guaranteeing safety; rather we focus on the issue of learning/modeling trajectory uncertainty, which such mechanisms must leverage for their safety guarantees.

\section{EXPERIMENT SETUP}
\label{sec:training}

The remainder of this paper aims to highlight the limitations of the aforementioned uncertainty models when considering human behavior. We show that the prevalent model classes of uncertainty (see Table 1) fail to capture human behavior at safety-critical thresholds ($\delta \leq 10^{-8}$), and exhibit significant errors when evaluated against real-world data. In particular, we test these uncertainty models on real-world driving data from the highD dataset \cite{highDdataset}, which uses overhead drones to capture vehicle trajectories from human drivers on German highways. % Since safe planning algorithms assume their computed uncertainty distributions are accurate, significant errors invalidate any claimed safety guarantees. 

In this section, we detail how we processed the highD dataset to extract important features in order to train/test the different uncertainty models. In the following section, we evaluate the accuracy of these models.  

% In particular, we illustrate these limitations by testing prevalent uncertainty models on real-world driving data from the highD driving dataset \cite{highDdataset}, which captures vehicles driving on German highways. We train the different models in Table \ref{table:models} to capture the distribution over trajectories in a training set, and observe how well they capture the distribution over trajectories in a separate test set. 

\begin{figure*}[h!]
\centering
\includegraphics[width=0.62\textwidth]{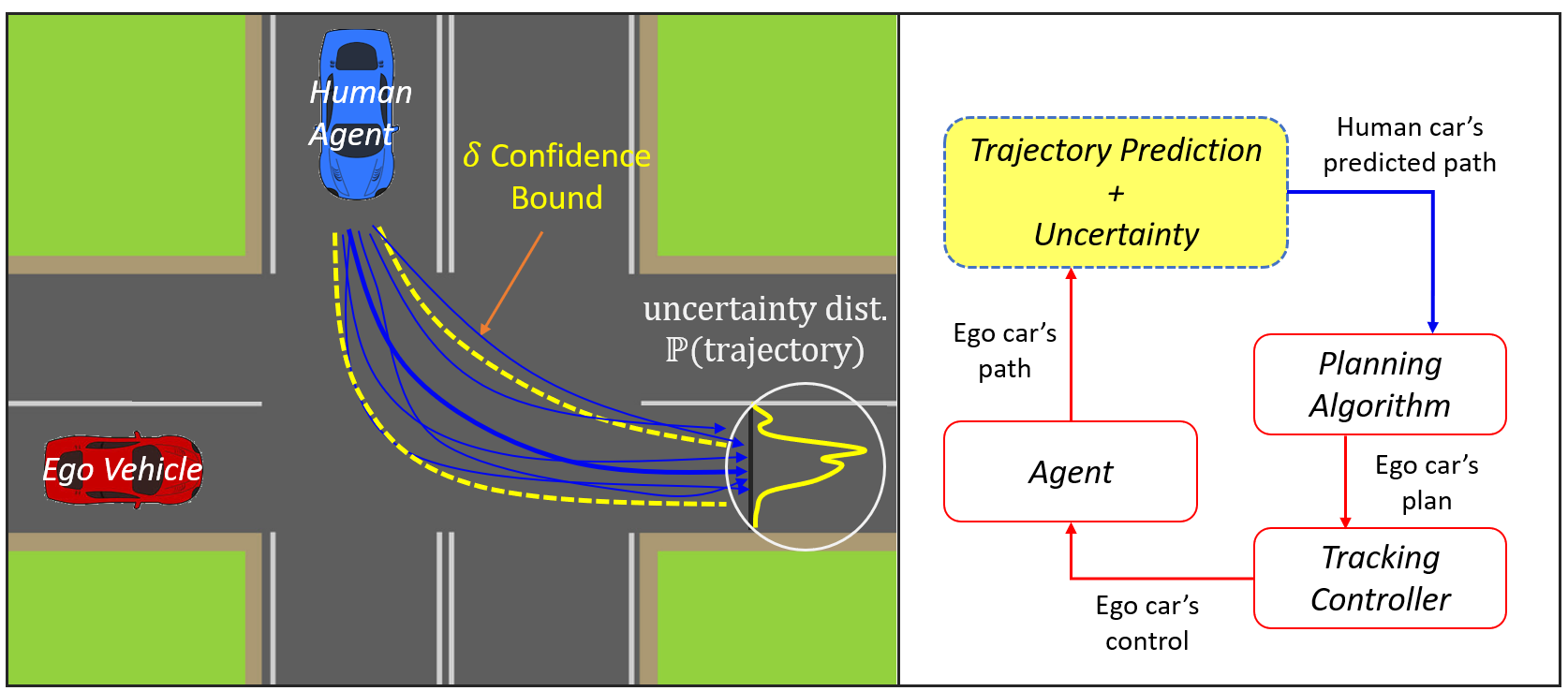}
\caption{\textbf{(Left)} In this example the red car must take into account the blue car's trajectory -- and its uncertainty -- in its plan to progress safely through the intersection. The dashed yellow curves denote the boundary of a tube that defines the $\delta$ confidence bound over trajectories. The white circle depicts a distribution over trajectories. The blue lines are example trajectories. \textbf{(Right)} Simplified illustration of different stages of the control pipeline. While every stage (prediction, planning, tracking) is crucial to guaranteeing safety, this paper focuses exclusively on the yellow box, \textit{prediction}. 
%Specifically, we analyze the ability of current approaches to learn $\delta$ confidence bounds, as any safety guarantees provided by the planning/control layers rely on accurate prediction/uncertainty quantification. 
}
\label{fig:scenario_illustration}
\end{figure*}

\subsection{Processing Dataset}

From the highD dataset, we extract all trajectories of length $10$ seconds, $\tau_{[0,10]}$ (denoting the agent's position over a $10$ second horizon), as well as the corresponding environmental context, $\mathcal{E}_{\tau}$, denoting the presence and position/velocity of surrounding cars. The trajectory and its context are denoted by the tuple $(\tau_{[0,10]}, \mathcal{E}_{\tau})$. We then split the trajectories/context into a training set, $\mathcal{D}^{train}$, and a test set, $\mathcal{D}^{test}$. For a given test trajectory, $(\tau^{(test)}_{[0:10]}, \mathcal{E}^{(test)}_{\tau}) \in \mathcal{D}^{test}$, we define \textit{equivalent scenarios}, $\mathcal{M}(\tau^{(test)}_{[0:10]}, \mathcal{E}^{(test)}_{\tau})$ as the set of trajectories with similar environmental context that are $\epsilon$-close ($\epsilon = 2$ft) over their first $2s$.

% along with their corresponding environmental context, $\mathcal{E}_{\tau}$ (i.e. position/velocity of surrounding cars) from the dataset. We then split the trajectories into a training set, $(\tau^{(train)}_{[0,10]}, \mathcal{E}^{(train)}_{\tau}) \in \mathcal{D}^{train}$, and test set, $(\tau^{(test)}_{[0:10]}, \mathcal{E}^{(test)}_{\tau}) \in \mathcal{D}^{test}$. For every test trajectory, $(\tau^{(test)}_{[0:10]}, \mathcal{E}^{(test)}_{\tau}) \in \mathcal{D}^{test}$, we collect all trajectories in the training set in equivalent scenarios,

\begin{equation}
\begin{split}
& \mathcal{M}(\tau^{(test)}_{[0:10]}, \mathcal{E}^{(test)}_{\tau}) = \Big\{ \tau_{[0:10]} ~ \Big| ~ (\tau_{[0:10]}, \mathcal{E}_{\tau}) \in \mathcal{D}^{train} ~ , \\ 
& ~~~~~~~~ \| \tau_{[0:2]} - \tau^{(test)}_{[0:2]} \|_{\infty} < \epsilon ~~ , ~~ \| \mathcal{E}_{\tau} - \mathcal{E}_{\tau}^{(test)} \|_{\infty} < \epsilon_{\mathcal{E}} \Big\} ~ .
\end{split}
\end{equation}

\begin{comment}
\begin{equation}
\begin{split}
& \mathcal{M}(\tau^{(test)}_{[0:10]}, \mathcal{E}^{(test)}_{\tau}) = \Big\{ \tau_{[0:10]} ~ \Big| ~ (\tau_{[0:10]}, \mathcal{E}_{\tau}) \in \mathcal{D}^{train} ~ , \\ 
& ~~~~~~~~~~~~~~~~~~~~~~~~~~~~~~~~ \| \tau_{[0:2]} - \tau^{(test)}_{[0:2]} \|_{\infty} < \epsilon ~ , \\
& ~~~~~~~~~~~~~~~~~~~~~~~~~~~~~~~~ \| \mathcal{E}_{\tau} - \mathcal{E}_{\tau}^{(test)} \|_{\infty} < \epsilon_{\mathcal{E}} \Big\} ~ .
\end{split}
\end{equation}
\end{comment}
% \note{The first sentence in the next paragraph is tough to follow.}

% We define \textit{equivalent scenarios} as the set of trajectories with similar environmental context that are $\epsilon$-close ($\epsilon = 2$ft) over their first $2s$. 
Therefore, $\mathcal{M}(\tau^{(test)}_{[0:10]}, \mathcal{E}_{\tau}^{(test)}) \subset \mathcal{D}^{train}$ denotes the set of scenarios within the training set, $\mathcal{D}^{train}$, that are highly similar to $(\tau^{(test)}_{[0:10]}, \mathcal{E}_{\tau}^{(test)})$. Since we have several test trajectories within $\mathcal{D}^{test}$, we define a pruned training set 
\begin{equation}
\mathcal{M}(\mathcal{D}^{test}) ~ = \bigcup_{(\tau, \mathcal{E}_{\tau}) \in \mathcal{D}^{test}} \mathcal{M}(\tau, \mathcal{E}_{\tau}) \subseteq \mathcal{D}^{train} ~.
\end{equation}
Every trajectory in the test set, $\mathcal{D}^{test}$, has equivalent scenarios in the pruned training set, $\mathcal{M}(\mathcal{D}^{test})$, such that we alleviate the issue of out-of-distribution error in learning. For clarity, let us define $\mathcal{T}^{train} = \mathcal{M}(\mathcal{D}^{test})$. % We emphasize that this helps us benchmark the best possible performance of the tested models.

% We chose a past observation horizon of $2s$ following \cite{Ding2019}, but found that the observed trends did not change considerably when using $1s$ or $3s$ for the past observation horizon. 

\subsection{Training Learned Uncertainty Models}

Given our test set, $\mathcal{D}^{test}$, and pruned training set, $\mathcal{T}^{train} \subseteq \mathcal{D}^{train}$, we would like to train a given uncertainty model $\hat{F}$ (e.g. Gaussian) on $\mathcal{T}^{train}$, and observe how accurately it captures the distribution of trajectories within $\mathcal{D}^{test}$. 

Let us divide a given scenario $(\tau_{[0,10]}, \mathcal{E}_{\tau})$ into the agent's state $x = (\tau_{[0:2]}, \mathcal{E}_{\tau})$, and its action $a = \tau_{[2:10]}$ (its future trajectory). Since the action is drawn from some unknown distribution over trajectories, $a \sim \mathcal{A}(x)$, our goal is to train a model $\hat{F}(x)$ that accurately approximates $\mathcal{A}(x)$, minimizing the following error, 
\begin{equation}
    L^{out} = \mathbb{E} \big[ m \big( \mathcal{A}(x) \| \hat{F}(x) \big) \big] ~,
    %L_{out} = \mathbb{E}_{(x, \mathcal{A}) \sim D^{true}} \big[ \mathcal{D}_{TV} \big( \mathcal{A} \| \hat{F}(x) \big) \big]
    \label{eq:l_out}
\end{equation}
where $m$ defines some metric over probability distributions. Clearly we do not know the true distribution $\mathcal{A}(x)$, but we can obtain an empirical estimate based on any dataset $\mathcal{D}$. We denote this empirical estimate $\hat{\mathcal{A}}(x, \mathcal{D})$. 

Using our pruned training dataset, $\mathcal{T}^{train}$,  we can train our uncertainty model $\hat{F}(x)$ (e.g. Gaussian, quantile, etc.), to minimize the following error function:
\begin{equation}
    L^{train} = \sum_{x \in \mathcal{T}^{train}} \big[ m \big( \hat{\mathcal{A}}(x, \mathcal{T}^{train}) ~ \| ~ \hat{F}(x) \big) \big] ~ .
\end{equation}
Then we can test the uncertainty model $\hat{F}(x)$ on the test dataset $\mathcal{D}^{test}$, yielding the error function
\begin{equation}
    L^{test}_{seen} = \sum_{x \in \mathcal{D}^{test}} \big[ m \big( \hat{\mathcal{A}}(x, \mathcal{D}^{test})) ~ \| ~ \hat{F}(x) \big) \big] ~ .
\end{equation}

Note that the pruned training set $\mathcal{T}^{train}$ contains data from all states $x$ represented in the test set $\mathcal{D}^{test}$. This alleviates issues associated with out-of-distribution data, such that $L^{test}_{seen}$ captures aleatoric uncertainty (vs. epistemic uncertainty). Because we do not have to consider generalization of our models to unseen (out-of-distribution) states, the following relationship generally holds,
\begin{equation}
L^{out}\underbrace{\geq}_{\text{generalization gap}} L^{test}_{seen}.
\end{equation}
In our analysis, we focus on $L^{test}_{seen}$ when measuring performance of our model $\hat{F}$. As this ignores generalization gap (how out-of-distribution examples affect model accuracy), it benchmarks \textit{best potential performance} of each model class.

\vskip 0.06 true in %\medskip
\noindent
\textbf{Accounting for replanning:} Most motion planning algorithms re-plan their trajectory at some fixed frequency (e.g. 1Hz). To account for this, we examine prediction error (e.g. violation of the $\delta-$uncertainty bound) only within a short re-planning horizon. I.e. the prediction must be accurate only within this replanning horizon. The horizon is set to $2$ sec.

\vskip 0.06 true in %\medskip
\noindent
\textbf{Incorporating conservative assumptions:} To further highlight the fundamental limitations of learning uncertainty models of human behavior, since many prediction algorithms leverage goal inference, we assume that an oracle gives us the target lane of every trajectory. Note that our aim is to illustrate limitations of learned probabilistic models, even under ideal conditions. Thus, this strong assumption (though unrealistic) helps us reason about the best-case scenario for each model class, providing an upper-bound on performance.

% \textbf{Accounting for multi-agent interactions:} A valid criticism of simply looking at the distributional error $m \big( \hat{\mathcal{A}}(x) \| F(x) \big)$ is that our distribution only depends on each individual agent's state, and not the multi-agent context, which clearly influences the trajectory distribution (e.g. the blue car impacts the red car's trajectory in Fig. \ref{fig:scenario_illustration}). However, the influence of other agents is typically reflected through the endpoint of our trajectory, $\tau$ (e.g. the influence of an agent cutting into our lane is reflected in our rapid slowdown). Therefore, we can account for multi-agent interactions by assuming that an oracle gives us \textit{the exact endpoint of every trajectory} $\tau_{[10]}$. While this is a generous assumption, it allows us to conservatively deal with the influence of other agents, under the assumption that the influence of other agents is reflected in the endpoint of our trajectory, $\tau_{[10]}$. % (as well as any mixture models) 

\vskip 0.05 true in
\noindent
Summarizing, we consider \textit{(a)} there is no generalization gap, and \textit{(b)} we are given the target lane of every trajectory. % Note that our objective is to illustrate limitations of learned probabilistic models, even under ideal conditions. Therefore, our strong assumptions (though unrealistic) help us reason about the best-case scenario for each model class, providing an upper-bound on performance. % (i.e. accuracy). % \textit{(c)} the model $\hat{F}$ optimally fits the training data (under the given model class), and \textit{(d)} we are given the end point of every trajectory (perfectly inferring our agent's goal and the influence of other agents). Utilizing these assumptions allows us to reason about the best-case scenario for each model class, providing an upper-bound on its performance (i.e. accuracy).
\vspace{0.3em}
\begin{quotation}
\noindent
\textit{If the models perform poorly under these extremely generous assumptions, we can not expect reasonable performance in realistic settings.}
\end{quotation}
% In the following subsections, we explore the limitations of different uncertainty model classes.

% In Section \ref{modeling}, we propose a different framework for dealing with human uncertainty in safety-critical applications, and in Section \ref{safety}, we show how safety can be guaranteed under this framework.

\section{RESULTS - ERROR IN UNCERTAINTY MODELS}
\label{sec:results}

In this section, we analyze the accuracy of different uncertainty models in capturing the distribution of trajectories in $\mathcal{D}^{test}$, after being trained on $\mathcal{T}^{train}$.

\subsection{Gaussian Uncertainty Models} \label{gaussian}

We start by analyzing the popular Gaussian uncertainty model, used in most UQ neural networks \cite{Michelmore2019}, Gaussian process models \cite{Aoude2013}, and robust regression \cite{liu2019,Nakka2020}. These approaches model the data and its uncertainty with a Gaussian distribution (see top 3 rows in Table 1). Using the procedure outlined in Section \ref{sec:training}, we compute the best-fit Gaussian distribution, $\hat{F}$, over the training trajectories $\mathcal{T}^{train}$, and observe how well it captures the in-distribution test trajectories in $\mathcal{D}^{test}$. 

\begin{figure}[h!]
\vskip -0.06 true in
\centering
\includegraphics[width=0.42\textwidth]{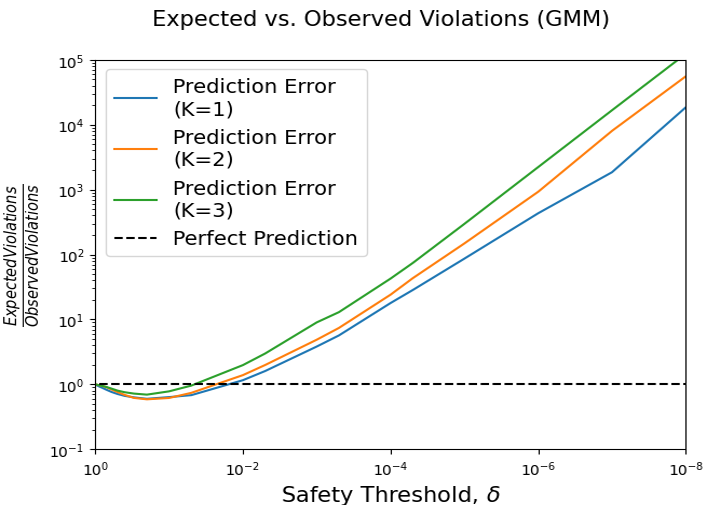}
\caption{Prediction error vs. safety threshold, $\delta$, using Gaussian mixture models on the highD dataset, considering a $2s$ re-planning horizon. $K$ denotes the number of mixtures used, with $K=1$ denoting a standard Gaussian distribution. The dashed black line represents a perfect prediction model.}
\label{fig:GMM_results}
\vskip -0.1 true in
\end{figure}

Figure \ref{fig:GMM_results} $(K=1)$ shows the ratio of observed to expected violations in the test set at each safety threshold, $\delta$. A \textit{violation} is defined when the test trajectory lies outside the $\delta$-uncertainty bound predicted by $\hat{F}$ (within a $2$s re-planning horizon) for a specified $\delta$. If the data followed a perfect Gaussian distribution, each curve in Fig. \ref{fig:GMM_results} would track the dotted black line (i.e. ratio near 1). If the curve falls below the dotted black line, then the model is overly conservative, and vice versa. We see that while the Gaussian model might be valid for $\delta \geq 0.01$, it is highly inaccurate outside this range, posing a problem for safety-critical applications.

\vskip 0.06 true in %\medskip
\noindent
\textbf{Gaussian mixture models (GMM): } One might point out that problems with the Gaussian model could be alleviated using GMMs over a discrete set of goals (e.g. left versus right turn). For example, interacting Gaussian processes (IGP) leverage this tool to alleviate the freezing robot problem \cite{Trautman2015}. However, when we trained GMMs on the same data with different numbers of mixtures ($K=2,...,4$), prediction performance on test data did not improve for low $\delta$ (see Fig. \ref{fig:GMM_results}). These results illustrate limitations of any Gaussian-based uncertainty model (IGP, GMM, etc.), by highlighting that human behavioral variation is inherently non-Gaussian.

% \note{Does this "Accounting for replanning" paragraph go here?  It seems like it should be placed in the previous section that discusses your methodology}

% \vskip 0.06 true in %\medskip
% \noindent
% \textbf{Accounting for replanning:} Most motion planning algorithms re-plan their trajectory at some fixed frequency (e.g. 1Hz). To account for this, we consider a violation \textit{only if the agent leaves the $\delta-$uncertainty bound within a short re-planning horizon} (i.e. the prediction must only be accurate within this replanning horizon). The horizon is set to $2$ sec.

\begin{figure}[!h]
\centering
\includegraphics[width=0.46\textwidth]{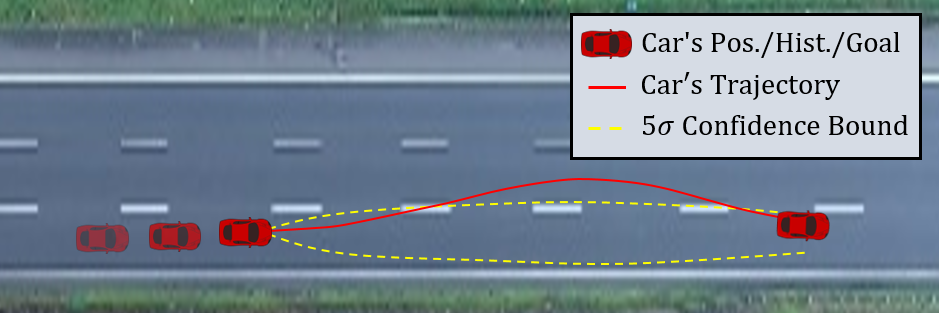}
\caption{Example of a car's trajectory, along with the approximate $5\sigma$ confidence bound computed from the training trajectories, given the car's target lane 8 seconds in the future.}
\label{fig:gaussian_constrain}
\end{figure}

\vskip 0.06 true in %\medskip
In addition to the issue of inaccurate distributional assumptions, the confidence bounds at level $\delta \approx 10^{-8}$ become very large, making planning around these bounds difficult or potentially infeasible. Figure \ref{fig:gaussian_constrain} shows the $5\sigma$ confidence tube projecting the position of a car forward in time, based on the trained model $\hat{F}$. The $5\sigma$ tube (corresponding to $\delta \approx 10^{-7}$) encroaches on each lane, making it difficult for other cars to drive alongside it. This is because, although the car will typically stay in its lane, in rare instances (see Figure \ref{fig:gaussian_constrain}) it will unexpectedly swerve into the other lane. This illustrates the difficulty of balancing the safety-efficiency tradeoff, as accounting for rare events may be necessary for safety-critical applications but introduces significant conservatism. % The $2\sigma$ ellipsoid provides a reasonable description for the future uncertain evolution of the red car. However, the $6\sigma$ ellipsoid (corresponding to $\delta \approx 10^{-9}$) encroaches on each lane, making it difficult for other cars to drive alongside it. % This is the well-known \textit{freezing robot problem} (FRP).

To further emphasize fragility of the Gaussian model at low $\delta$, we generated synthetic 2D data from different, known distributions, and examined how well the best fit Gaussian predicted violations at a given $\delta$. Even with perfectly i.i.d. training/test data, the error at low $\delta$ was significant. Details and results are in Appendix A (found at\cite{full_report}).
% To further emphasize fragility of the Gaussian model at low $\delta$, we generated synthetic 2D data following 3 different, known distributions. We then examined how well the best fit Gaussian predicted violations at a given $\delta$. Even with perfectly i.i.d. training/test data, the error at low $\delta$ was significant (off by an order-of-magnitude) for non-Gaussian distributions. See Appendix A for details.

\subsection{Noisy Rational Model}

% The noisy rational model has become increasingly popular in recent years as a model of human behavior and uncertainty; it assumes 
The noisy rational model considers that humans behave approximately optimally with respect to some reward function. It has enabled significant progress in inverse reinforcement learning (IRL) by allowing researchers to learn reward functions from human data \cite{Sadigh2016}, and compute explicit uncertainty intervals over human agents' actions \cite{FridovichKeil2020}. However, the noisy rational model adopts an underlying model of uncertainty in the exponential family, which places a strong assumption on the shape of the uncertainty distribution and assumes that there is a single “optimal” trajectory:
\begin{equation}
\mathbb{P}(x_{t+1} ~ | ~ \beta) = \frac{e^{\beta Q_H(x_{t+1})}}{\sum_{\tilde{x}_{t+1}} e^{\beta Q_H(\tilde{x}_{t+1})}} .
\label{eq:noisy_rational}
\end{equation}
In our driving scenario, the optimal model simplifies to the Gaussian distribution, since $Q_H = \| x_{t+1} - \hat{x}_{t+1} \|_{\Sigma}$ for some $\Sigma$ (i.e. we want to best fit the data). As a result, the issues illustrated in Figures \ref{fig:GMM_results} and \ref{fig:gaussian_constrain} are exactly faced by the noisy rational model (i.e. the shape of the underlying distribution does not match the assumed distribution). Thus, even in the best case -- known target lane, no generalization gap -- these models are ill-equipped to provide safety guarantees for safety-critical systems ($\delta < 10^{-8}$). % Recall that these results compute the best Gaussian fit after observing all the training data, which upper-bounds the performance of any approach leveraging this uncertainty model.

\subsection{Quantile Regression}

Quantile regression is an appealing alternative as it does not require strong assumptions on the underlying uncertainty distribution \cite{Fan2020}. It is only concerned with computing tubes such that $1-\delta$ proportion of trajectories are within that tube and $\delta$ are outside. To demonstrate its performance, we again use the procedure outlined in Section \ref{sec:training} to train a quantile regression model $\hat{F}$. The quantile bounds are approximated as the smallest convex tube containing $1-\delta$ proportion of trajectories, which optimizes the expected mutual information between the state, $x$, and action, $a$ \cite{Pelckmans2008}. 

\begin{figure}[h!]
    \centering
    \includegraphics[width=0.45\textwidth]{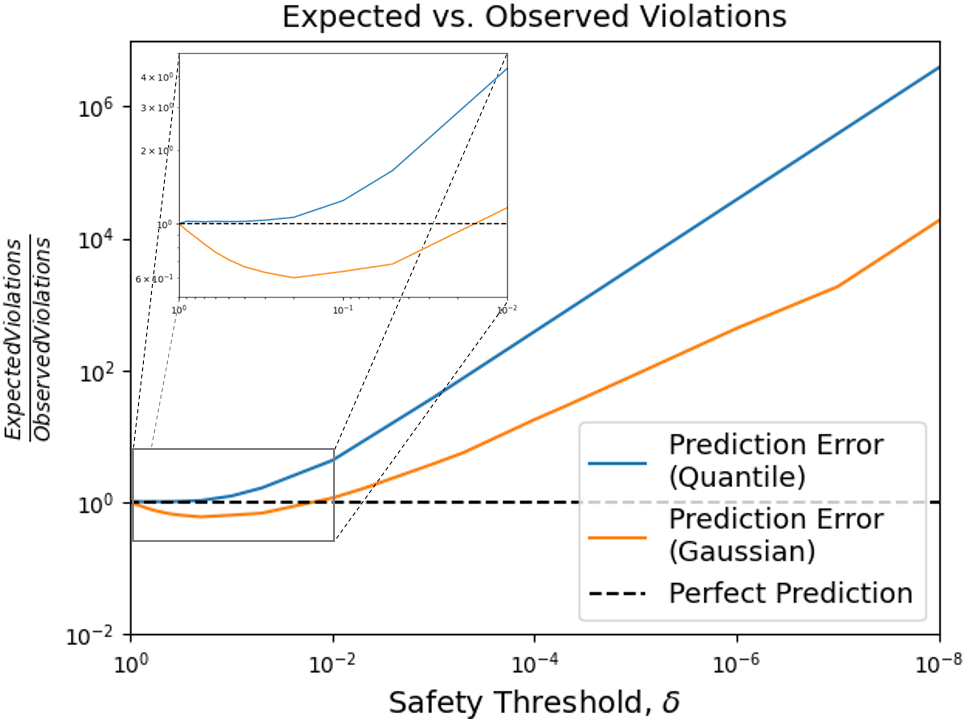}
    \caption{Prediction error vs. safety threshold, $\delta$ using computed quantile bounds or Gaussian uncertainty model on the highD dataset (assuming $2s$ re-planning horizon). The dashed black line represents a perfect prediction model.}
    \label{fig:quantile}
\end{figure}

Figure \ref{fig:quantile} shows the ratio of the observed to expected number of test trajectories outside each quantile at safety threshold $\delta$. As seen in the plot, the quantile regression model performs much better than the Gaussian model for $\delta > 0.1$. However, performance rapidly deteriorates as $\delta$ decreases, making estimated confidence bounds meaningless, since they fail to predict violation probabilities.  % (which makes sense as it does not place strong assumptions on the shape of the uncertainty distribution), and it is quite accurate in the range $\delta > 0.05$. However, performance rapidly deteriorates as $\delta$ decreases, making the estimated confidence bounds meaningless. 

This result makes sense, as obtaining accurate quantile bounds at the $\delta$-confidence level relies on splitting the data: $\delta$ percent of points should be outside the quantile bound with the rest inside those bounds. However, little (if any) data is available outside the quantile bound for very low $\delta$. Put differently, to \textit{observe} a one-in-a-million event, we would need to see a million trajectories. To \textit{reliably predict} those events, we would need many more trajectories. % As a result, accurate quantile regression is still limited to a narrow confidence range.

% \note{This next paragraph(s) is important.  perhaps start the paragraph with a bold face title?}

\begin{figure*}[h!]
\centering
\includegraphics[width=0.76\textwidth]{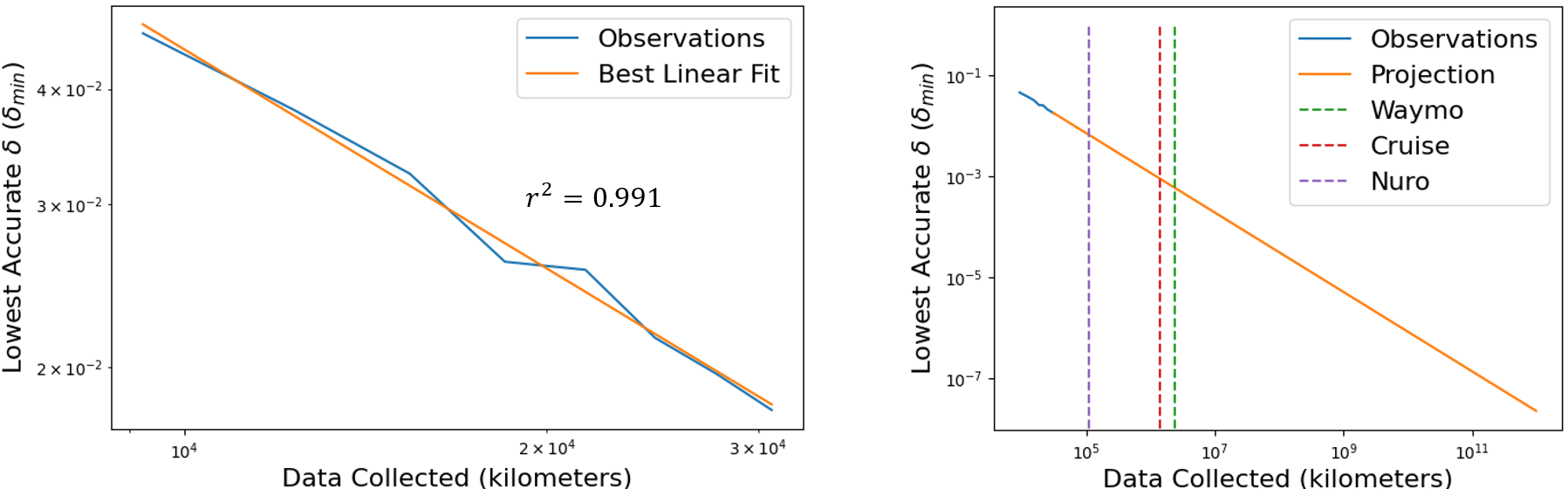}
\caption{\textit{(a)} Smallest accurate $\delta$ versus amount of data collected. The trend is highly linear ($r^2 \approx 0.995)$, and holds across different sections of the dataset. \textit{(b)} Projection showing the expected amount of data required to obtain an accurate safety threshold $\delta_{min}$. The dashed lines show the number of kilometers driven in California in 2019 by Waymo, Cruise, and Nuro.}
\label{fig:quantile_scaling}
\end{figure*}

\vskip 0.06 true in
\noindent
\textbf{Improving Accuracy with Increasing Data:} Given the availability of increasingly large robotics datasets, we should ask whether we could reach good accuracy at desired safety thresholds, $\delta$, by using more data. To answer this, we define the  smallest accurate safety threshold, $\delta_{min}$, as follows, % In Figure \ref{fig:quantile_linear}, we plot the smallest accurate safety threshold, $\delta_{min}$, versus the amount of data, where $\delta_{min}$ is defined as
\begin{equation}
    \delta_{min} ~ = ~ \min ~ \delta ~~~~~ \text{such that} ~~~~~  \left\lvert \log \Bigg( \frac{\text{expected}(\delta)}{\text{observed}(\delta)} \Bigg) \right\rvert \leq \varepsilon ~ .
\end{equation}
We set $\varepsilon = 0.5$, where $\varepsilon$ represents the vertical distance between each curve in Fig \ref{fig:quantile} and the dotted black line. Thus, $\delta_{min}$ represents the smallest $\delta$ such that our computed quantile bounds are $\varepsilon$-accurate. Note that $\delta_{min}$ is computed with respect to a given set of data. Therefore, by varying the size of our training set, we capture how $\delta_{min}$ varies with the amount of training data, shown in Fig. \ref{fig:quantile_scaling}a. The trend shown in Fig. \ref{fig:quantile_scaling}a is surprisingly linear ($r^2 = 0.995$), which held across different sections of the dataset (i.e. different highways). This scaling is consistent with the lower bound on sample complexity derived in \cite{Ehrenfeucht1989}, shown in Eq. \eqref{eq:scaling} and discussed further below. % , suggesting that the quantile regression results shown are close to optimal (w.r.t. sample complexity). This lower bound is shown in Equation \eqref{eq:scaling} and discussed further below. % We found a surprisingly strong linear trend between $\log{(\text{minimum}~\delta)}$ and $\log{(\text{amount of data})}$, which held across different datasets, scenarios, and train/test splits.

% \note{In 2018, it was estimated that 3.225 trillion miles were driven in the U.S. by all cars/trucks.  So, your 10 trillion numbers is data from every car in the U.S. for 2 years.}

While initially promising, if we project this linear trend down to $\delta_{min} \approx 10^{-8}$, we find that the amount of data required to reach safety-critical thresholds is far from feasible. Figure \ref{fig:quantile_scaling}b shows that we would need trillions of kilometers of driving data to achieve accurate quantile bounds, even under extremely generous assumptions (e.g. perfect generalization). For reference, in 2018, approximately 5 trillion kilometers were driven in total across all cars/trucks in the U.S. \cite{nhtsa}. 
\begin{comment}
\vspace{-0.25em}
\begin{center}
    \textit{Even assuming perfect generalization from one driving scenario to another, we would need almost a million times more driving data than the to-date cumulative driving kilometers of Waymo.}
\end{center}
\vspace{-0.25em}
\end{comment}

We conducted the same analysis on synthetic 2D data, and found the same trends seen in Figures \ref{fig:quantile} and \ref{fig:quantile_scaling}. Details and results are in Appendix B (found at \cite{full_report}).

\vskip 0.06 true in
\noindent
\textbf{Quantile Regression as a Fundamental Limitation:} One might be tempted to conclude from Figure \ref{fig:quantile_scaling}a that we should look for alternative methods with better data efficiency. % (i.e. allow us to accurately infer confidence bounds at smaller $\delta$ with less data).
However, all methods providing confidence bounds over trajectories at some safety threshold $\delta$ can be fundamentally viewed as classification problems; we must classify $1 - \delta$ trajectories within some learned bounds, with the rest outside those bounds. By viewing this as a classification problem, we can leverage results from VC-analysis that lower bound the data required, $N$, to guarantee a given prediction confidence \cite{Ehrenfeucht1989}: To guarantee $\textit{Pr}(\textnormal{error}) \leq \delta$, we require
\begin{equation}
    N(\delta, M) = \Omega \Big ( \frac{1}{\delta} \ln \big( \frac{1}{\delta} \big) + \frac{\text{VCdim}(M)}{\delta} \Big)
    \label{eq:scaling}
\end{equation}
where VCdim(M) is the VC dimension of the utilized model $M$ (see \cite{Ehrenfeucht1989} for proof). The linear trend in Fig. \ref{fig:quantile_scaling} (showing $N(\delta) \propto \frac{1}{\delta_{min}}$) fits very nicely with the lower bound \eqref{eq:scaling}, given that the second term dominates the first (i.e. we have large VC dimension). Note that if the first term dominated the second term, we would expect worse data scaling. % This suggests that alternative methods cannot provide confidence bounds with better data scaling than shown in Figure \ref{fig:quantile_scaling}. 
\begin{center}
    \textit{This suggests alternate methods cannot provide confidence bounds with better data scaling than shown in Figure \ref{fig:quantile_scaling}.}
\end{center}
\begin{comment}
\vspace{0.3em}
\begin{quotation}
\noindent
\textit{This analysis suggests that alternative methods cannot provide confidence bounds with better data scaling than shown in Figure \ref{fig:quantile_scaling}.}
\end{quotation}
\end{comment}
% \note{The last sentence of the paragraph immediately above is important.  Is there some way for us to "highlight it". Perhaps in italics, or make it a "remark."} 
\subsection{Other Uncertainty Models}

Due to space constraints, we discuss remaining models in Appendix C (found at\cite{full_report}), including: \textit{(1)} generative models, \textit{(2)} scenario optimization, and \textit{(3)} hidden Markov models (HMM). However, below, we briefly describe fundamental problems each of these models face:
\begin{itemize}[leftmargin=*]
\item Scenario optimization -- similar to quantile regression -- requires far too much data to be feasible for small $\delta$. Even with 40,000 trajectories in equivalent scenarios, we only reach $\delta \approx 10^{-4}$. 
\item Generative models \textit{implicitly} learn the distribution $\mathcal{A}(x) = p(a | x)$. While promising, it has been shown -- empirically and theoretically -- that they can fail to learn the true distribution, even when their training objective nears optimality \cite{arora2018}. Also, using state-of-the-art models \cite{Gupta2018,Sadeghian2019}, it would require at least a day to generate enough trajectories to certify safety with sufficiently high confidence. % $\delta \approx 10^{-8}$. %, making online feasibility unlikely. % Importantly, it is not yet clear how the implicit distribution representation can be used to construct explicit safety guarantees. % ; one could imagine this being done by generating billions/trillions of random trajectories, though the computational cost would quickly become infeasible. See Appendix for further discussion. 
\item HMMs are distinct as they learn probabilities over discrete states (e.g. goal positions). However, in the Appendix we show that even with a known observation function,  $\mathbb{P}(\textnormal{obs} | \textnormal{state})$, it is highly unlikely to obtain sufficient confidence ($\delta \approx 10^{-8}$) of being in a given state.
\end{itemize}

% \note{In the paragraph immediately above, you say: "However, we show that even with a known observation function.....".  How actually do you show that--did some appendix or other block of text get taken out (that did the actually "showing"). Right now the text doesn't support this "we show" aspect}

\vskip 0.06 true in
\noindent
\textbf{Note on UQ Neural Networks:} We have not discussed UQ neural networks, because neural networks do not compose a distinct class of uncertainty models. Instead, they only provide a functional representation of the uncertainty in a given class (e.g. UQ neural networks typically output a Gaussian distribution). Our results highlight \textit{best-case} performance bounds for each class of model uncertainties, given optimal fit to the data. Thus, using neural networks to parameterize the model uncertainty will only yield worse performance.

% \smallskip
% \textit{Code for all results in the paper and Appendix can be found in the supplementary materials.}

\section{CONCLUSION AND FUTURE WORK}

Our main message is that even under extremely generous assumptions, current models of human uncertainty are unable to extend safety guarantees to the confidence levels, e.g. $\delta < 10^{-8}$, that are needed for widespread adoption of safety-critical autonomy in human environments. 
\begin{center}
    \textit{Learned uncertainty distributions become highly inaccurate at low $\delta$, undermining any claimed guarantees of safety.}
\end{center}
There is a fundamental limitation to modeling human uncertainty purely as a random process. Data-driven methods (i.e. machine learning) are designed to capture prominent patterns in data not to predict rare events. Intuitively, we need a million samples just to \textit{observe} a one-in-a-million event. While it is possible that huge datasets could eventually enable accurate prediction of rare events, our analysis shows that such amounts of data are infeasible in the near future. %(even ignoring generalization issues and computational cost). % As shown in Fig. \ref{fig:quantile_scaling}, one potential solution is to collect more data. While data requirements are huge, more than 3 trillion miles are driven per year in the US. Collecting and utilizing this amount of data would pose new challenges and require concerted effort, but could be a potential solution. 
%The issues described above illustrate a fundamental limitation of modeling human uncertainty purely as random processes.

% If it is infeasible to learn uncertainty bounds at probability levels $\delta = 10^{-8}$, then how can we guarantee safety of systems at these levels?
\vskip 0.06 true in
\noindent
\textbf{Human uncertainty vs. sensor-based uncertainties:} Even if a system must be certified safe with $\delta=10^{-8}$, it is uncommon to require any single module to have a failure probability less than $10^{-8}$. Instead, redundancy with multiple, independent modules can help certify system safety. The key is that the modules must be independent. While this may be a fair assumption for sensing uncertainty, it is not fair for human behavior prediction. % However, this could be a promising avenue of research: to simultaneously learn multiple predictors \cite{Filos2020}, while enforcing independence between their predictions. % For example, in a robotic system, two different modules (one using LIDAR and one using stereo cameras) might predict an obstacle's current position, each with confidence $1 - 10^{-4}$. Then the overall system can reason about the obstacle's position with confidence $1 - 10^{-8}$. The key is that \textit{the modules must be independent}. While this may be a fair assumption for sensing uncertainty, it is not fair for human behavior prediction. However, this could be a promising avenue of research: to simultaneously learn multiple predictors \cite{Filos2020}, \textit{while enforcing independence between their predictions}. % No sensor exists to measure a human agent's \textit{future} position, and it is unclear how to enforce independence of multiple predictive modules. However, this is an interesting avenue of research: to simultaneously learn multiple predictors \cite{Filos2020}, \textit{while enforcing independence between their predictions}. % This is a unique challenge of human uncertainty. However, one potential solution may be to simultaneously learn ensemble predictors, \textit{while enforcing/verifying independence between their predictions} \cite{}.

%\smallskip
\vskip 0.06 true in
% \note{Add a reference after the italicized "assume-guarantee contracts"}
\noindent
\textbf{Future Work:} A promising solution to guarantee safety at low $\delta$ is to utilize prior knowledge about human behavior; in particular, humans obey interaction rules (e.g. signaling intent) \cite{Bratman1992}, which bound uncertainty in useful ways and can be encoded in \textit{assume-guarantee contracts} \cite{phan2019}.
\begin{comment}
\begin{definition}{\cite{phan2019}} An assume-guarantee contract for an agent is a 2-tuple $(\mathcal{A},\mathcal{G})$ where, 
\vspace{-0.35em}
\begin{itemize}
    \item $\mathcal{A}$ is a set of behavioral constraints that the agent assumes its environment to have.
    \item $\mathcal{G}$ is  a  set  of  behavioral constraints that it must obey, as long as its environment satisfies $\mathcal{A}$.
\end{itemize}
\end{definition} 
\vspace{-0.35em}
\end{comment}
A contract might encode that an agent cannot mislead others about its intention, assuming that others do not mislead it. For example, an agent cannot first pretend to yield to a merging vehicle, before speeding up to hit it. We thus propose trading one challenge for another: rather than learning uncertainty bounds that agents obey with probability $1-\delta$, we should aim to specify interpretable contracts (i.e. behavioral constraints) with learned components that agents must surely obey. 

We believe such a framework is necessary to move away from treating uncertainty in human behavior purely as a random process. Instead, human uncertainty can be constrained by combining \textit{learned components} that predict expected actions and \textit{prior knowledge} restricting the danger of rare events in a rigorous, interpretable manner.

% Such a framework is described in \cite{Cheng2021}, which We believe such a framework is necessary to move away from treating uncertainty in human behavior purely as a random process. Instead, human uncertainty can be constrained by combining \textit{learned components} that predict expected actions and \textit{prior knowledge} restricting the danger of rare events in a rigorous, interpretable manner. % However, regardless of the framework used, our results highlight that current probabilistic guarantees from learned uncertainty models are not accurate nor adequate for safety-critical applications.


%%%%%%%%%%%%%%%%%%%%%%%%%%%%%%%%%%%%%%%%%%%%%%%%%%%%%%%%%%%%%%%%%%%%%%%%%%%%%%%%

%%%%%%%%%%%%%%%%%%%%%%%%%%%%%%%%%%%%%%%%%%%%%%%%%%%%%%%%%%%%%%%%%%%%%%%%%%%%%%%%

%%%%%%%%%%%%%%%%%%%%%%%%%%%%%%%%%%%%%%%%%%%%%%%%%%%%%%%%%%%%%%%%%%%%%%%%%%%%%%%%
% \note{There are several messy or incomplete references}

\bibliographystyle{IEEEtran}
\bibliography{references}

\begin{thebibliography}{10}
\providecommand{\url}[1]{#1}
\csname url@rmstyle\endcsname
\providecommand{\newblock}{\relax}
\providecommand{\bibinfo}[2]{#2}
\providecommand\BIBentrySTDinterwordspacing{\spaceskip=0pt\relax}
\providecommand\BIBentryALTinterwordstretchfactor{4}
\providecommand\BIBentryALTinterwordspacing{\spaceskip=\fontdimen2\font plus
\BIBentryALTinterwordstretchfactor\fontdimen3\font minus
  \fontdimen4\font\relax}
\providecommand\BIBforeignlanguage[2]{{%
\expandafter\ifx\csname l@#1\endcsname\relax
\typeout{** WARNING: IEEEtran.bst: No hyphenation pattern has been}%
\typeout{** loaded for the language `#1'. Using the pattern for}%
\typeout{** the default language instead.}%
\else
\language=\csname l@#1\endcsname
\fi
#2}}

\bibitem{Fisac2018}
J.~Fisac, A.~Bajcsy, S.~Herbert, D.~Fridovich-Keil, S.~Wang, C.~Tomlin, and
  A.~Dragan, ``{Probabilistically Safe Robot Planning with Confidence-Based
  Human Predictions},'' in \emph{Robotics: Science and Systems}, 2018.

\bibitem{Aoude2013}
G.~S. Aoude, B.~D. Luders, J.~M. Joseph, N.~Roy, and J.~P. How,
  ``{Probabilistically safe motion planning to avoid dynamic obstacles with
  uncertain motion patterns},'' \emph{Autonomous Robots}, vol.~35, no.~1, pp.
  51--76, 2013.

\bibitem{FridovichKeil2020}
D.~Fridovich-Keil, A.~Bajcsy, J.~F. Fisac, S.~L. Herbert, S.~Wang, A.~D.
  Dragan, and C.~J. Tomlin, ``Confidence-aware motion prediction for real-time
  collision avoidance,'' \emph{International Journal of Robotics Research},
  vol.~39, no. 2-3, pp. 250--265, March 2020.

\bibitem{Nakka2020}
Y.~K. Nakka, A.~Liu, G.~Shi, A.~Anandkumar, Y.~Yue, and S.-J. Chung,
  ``{Chance-Constrained Trajectory Optimization for Safe Exploration and
  Learning of Nonlinear Systems},'' \emph{arXiv}, 2020.

\bibitem{Fan2020}
D.~D. Fan, A.~Agha-mohammadi, and E.~A. Theodorou, ``Deep learning tubes for
  tube mpc,'' \emph{arXiv}, 2020.

\bibitem{Shalev-Shwartz2017}
S.~Shalev-Shwartz, S.~Shammah, and A.~Shashua, ``{On a Formal Model of Safe and
  Scalable Self-driving Cars},'' \emph{arXiv}, 2017.

\bibitem{nhtsa}
``National highway traffic safety administration. traffic safety facts annual
  report,'' 2019.

\bibitem{highDdataset}
R.~Krajewski, J.~Bock, L.~Kloeker, and L.~Eckstein, ``The highd dataset: A
  drone dataset of naturalistic vehicle trajectories on german highways for
  validation of highly automated driving systems,'' in \emph{International
  Conference on Intelligent Transportation Systems (ITSC)}, 2018.

\bibitem{Gupta2018}
A.~Gupta, J.~Johnson, L.~Fei-Fei, S.~Savarese, and A.~Alahi, ``Social gan:
  Socially acceptable trajectories with generative adversarial networks,'' in
  \emph{Proceedings of the IEEE Conference on Computer Vision and Pattern
  Recognition (CVPR)}, June 2018.

\bibitem{Fisac2019}
J.~F. Fisac, E.~Bronstein, E.~Stefansson, D.~Sadigh, S.~S. Sastry, and A.~D.
  Dragan, ``{Hierarchical game-theoretic planning for autonomous vehicles},''
  in \emph{International Conference on Robotics and Automation (ICRA)}, 2019,
  pp. 9590--9596.

\bibitem{Hakobyan2020}
A.~Hakobyan and I.~Yang, ``{Learning-Based Distributionally Robust Motion
  Control with Gaussian Processes},'' \emph{arXiv}, 2020.

\bibitem{Cheng2020}
R.~Cheng, M.~J. Khojasteh, A.~D. Ames, and J.~W. Burdick, ``{Safe Multi-Agent
  Interaction through Robust Control Barrier Functions with Learned
  Uncertainties},'' \emph{arXiv}, 2020.

\bibitem{Xu2014}
W.~Xu, J.~Pan, J.~Wei, and J.~M. Dolan, ``{Motion planning under uncertainty
  for on-road autonomous driving},'' in \emph{IEEE ICRA}, 2014.

\bibitem{Sadigh2016a}
D.~Sadigh and A.~Kapoor, ``{Safe control under uncertainty with Probabilistic
  Signal Temporal Logic},'' in \emph{Robotics: Science and Systems}, vol.~12,
  2016.

\bibitem{forghani2016}
M.~Forghani, J.~M. McNew, D.~Hoehener, and D.~{Del Vecchio}, ``{Design of
  driver-assist systems under probabilistic safety specifications near stop
  signs},'' \emph{IEEE Transactions on Automation Science and Engineering},
  vol.~13, no.~1, pp. 43--53, 2016.

\bibitem{Michelmore2019}
R.~Michelmore, M.~Wicker, L.~Laurenti, L.~Cardelli, Y.~Gal, and M.~Kwiatkowska,
  ``{Uncertainty Quantification with Statistical Guarantees in End-to-End
  Autonomous Driving Control},'' in \emph{IEEE International Conference on
  Robotics and Automation (ICRA)}, 2020, pp. 7344--7350.

\bibitem{sadigh2014}
D.~Sadigh, K.~Driggs-Campbell, A.~Puggelli, W.~Li, V.~Shia, R.~Bajcsy, A.~L.
  Sangiovanni-Vincentelli, S.~S. Sastry, and S.~A. Seshia, ``{Data-driven
  probabilistic modeling and verification of human driver behavior},'' in
  \emph{AAAI Spring Symposium}, 2014, pp. 56--61.

\bibitem{Liu2015}
W.~Liu, S.~W. Kim, S.~Pendleton, and M.~H. Ang, ``{Situation-aware decision
  making for autonomous driving on urban road using online POMDP},'' in
  \emph{2015 IEEE Intelligent Vehicles Symposium (IV)}, 2015, pp. 1126--1133.

\bibitem{Cesari2017}
G.~{Cesari}, G.~{Schildbach}, A.~{Carvalho}, and F.~{Borrelli}, ``Scenario
  model predictive control for lane change assistance and autonomous driving on
  highways,'' \emph{IEEE Intelligent Transportation Systems Magazine}, vol.~9,
  no.~3, pp. 23--35, 2017.

\bibitem{Chen2020}
Y.~Chen, S.~Dathathri, T.~Phan-Minh, and R.~M. Murray, ``Counter-example guided
  learning of bounds on environment behavior,'' in \emph{Proceedings of the
  Conference on Robot Learning}, vol. 100.\hskip 1em plus 0.5em minus
  0.4em\relax PMLR, Nov 2020, pp. 898--909.

\bibitem{Sartipizadeh2020}
H.~Sartipizadeh and B.~Açıkmeşe, ``Approximate convex hull based scenario
  truncation for chance constrained trajectory optimization,''
  \emph{Automatica}, vol. 112, 2020.

\bibitem{Sadeghian2019}
A.~Sadeghian, V.~Kosaraju, A.~Sadeghian, N.~Hirose, H.~Rezatofighi, and
  S.~Savarese, ``{SoPhie: An attentive GAN for predicting paths compliant to
  social and physical constraints},'' in \emph{IEEE/CVF Conference on Computer
  Vision and Pattern Recognition (CVPR)}, 2019, pp. 1349--1358.

\bibitem{Salzmann2020}
T.~Salzmann, B.~Ivanovic, P.~Chakravarty, and M.~Pavone, ``{Trajectron++:
  Multi-Agent Generative Trajectory Forecasting With Heterogeneous Data for
  Control},'' \emph{arXiv}, 2020.

\bibitem{Ellis2009}
D.~{Ellis}, E.~{Sommerlade}, and I.~{Reid}, ``Modelling pedestrian trajectory
  patterns with gaussian processes,'' in \emph{IEEE 12th International
  Conference on Computer Vision Workshops, ICCV Workshops}, 2009, pp.
  1229--1234.

\bibitem{Trautman2015}
P.~Trautman, J.~Ma, R.~M. Murray, and A.~Krause, ``{Robot navigation in dense
  human crowds: Statistical models and experimental studies of human-robot
  cooperation},'' \emph{The International Journal of Robotics Research},
  vol.~34, no.~3, pp. 335--356, 2015.

\bibitem{ferguson2015}
S.~Ferguson, B.~Luders, R.~C. Grande, and J.~P. How, ``{Real-time predictive
  modeling and robust avoidance of pedestrians with uncertain, changing
  intentions},'' in \emph{Algorithmic Foundations of Robotics XI: Selected
  Contributions of the Eleventh International Workshop on the Algorithmic
  Foundations of Robotics}.\hskip 1em plus 0.5em minus 0.4em\relax Springer
  International Publishing, 2015, pp. 161--177.

\bibitem{liu2019}
A.~Liu, G.~Shi, S.-J. Chung, A.~Anandkumar, and Y.~Yue, ``{Robust Regression
  for Safe Exploration in Control},'' vol. 120.\hskip 1em plus 0.5em minus
  0.4em\relax PMLR, Jun 2020, pp. 608--619.

\bibitem{gray2013}
A.~Gray, Y.~Gao, T.~Lin, J.~K. Hedrick, and F.~Borrelli, ``{Stochastic
  predictive control for semi-autonomous vehicles with an uncertain driver
  model},'' in \emph{2013 IEEE Intelligent Vehicles Symposium (IV)}, 2013, pp.
  2329--2334.

\bibitem{Tagasovska2018}
N.~Tagasovska and D.~Lopez-Paz, ``{Single-Model Uncertainties for Deep
  Learning},'' in \emph{Neural Information Processing Systems}, 2018.

\bibitem{Campi2018}
M.~C. Campi and S.~Garatti, ``{Wait-and-judge scenario optimization},''
  \emph{Mathematical Programming}, 2018.

\bibitem{carvalho2015}
A.~Carvalho, S.~Lef{\'{e}}vre, G.~Schildbach, J.~Kong, and F.~Borrelli,
  ``{Automated driving: The role of forecasts and uncertainty - A control
  perspective},'' in \emph{European Journal of Control}, vol.~24, 2015, pp.
  14--32.

\bibitem{driggs-campbell2017}
K.~Driggs-Campbell, V.~Govindarajan, and R.~Bajcsy, ``{Integrating Intuitive
  Driver Models in Autonomous Planning for Interactive Maneuvers},'' \emph{IEEE
  Transactions on Intelligent Transportation Systems}, vol.~18, no.~12, pp.
  3461--3472, 2017.

\bibitem{driggs-campbell2018}
K.~Driggs-Campbell, R.~Dong, and R.~Bajcsy, ``{Robust, informative
  human-in-the-loop predictions via empirical reachable sets},'' \emph{IEEE
  Transactions on Intelligent Vehicles}, vol.~3, no.~3, pp. 300--309, 2018.

\bibitem{Li2016}
N.~Li, D.~Oyler, M.~Zhang, Y.~Yildiz, A.~Girard, and I.~Kolmanovsky,
  ``{Hierarchical reasoning game theory based approach for evaluation and
  testing of autonomous vehicle control systems},'' in \emph{IEEE Conference on
  Decision and Control}, 2016.

\bibitem{Sadigh2016}
D.~Sadigh, S.~Sastry, S.~A. Seshia, and A.~D. Dragan, ``{Planning for
  autonomous cars that leverage effects on human actions},'' in \emph{Robotics:
  Science and Systems}, 2016.

\bibitem{Kwon2020}
M.~Kwon, E.~Biyik, A.~Talati, K.~Bhasin, D.~P. Losey, and D.~Sadigh, ``When
  humans aren't optimal: Robots that collaborate with risk-aware humans,'' in
  \emph{Proceedings of the 2020 ACM/IEEE International Conference on
  Human-Robot Interaction}.\hskip 1em plus 0.5em minus 0.4em\relax Association
  for Computing Machinery, 2020, p. 43–52.

\bibitem{Kelley2008}
R.~Kelley, A.~Tavakkoli, C.~King, M.~Nicolescu, M.~Nicolescu, and G.~Bebis,
  ``Understanding human intentions via hidden markov models in autonomous
  mobile robots,'' in \emph{Proceedings of the 3rd ACM/IEEE International
  Conference on Human Robot Interaction}.\hskip 1em plus 0.5em minus
  0.4em\relax Association for Computing Machinery, 2008, p. 367–374.

\bibitem{McGhan2015}
C.~L. McGhan, A.~Nasir, and E.~M. Atkins, ``{Human intent prediction using
  Markov decision processes},'' \emph{Journal of Aerospace Information
  Systems}, 2015.

\bibitem{Bandyopadhyay2013}
T.~Bandyopadhyay, K.~S. Won, E.~Frazzoli, D.~Hsu, W.~S. Lee, and D.~Rus,
  ``{Intention-aware motion planning},'' in \emph{Springer Tracts in Advanced
  Robotics}, 2013.

\bibitem{Lam2014}
C.~P. Lam and S.~S. Sastry, ``{A POMDP framework for human-in-the-loop
  system},'' in \emph{IEEE Conference on Decision and Control}, 2014.

\bibitem{Tran2015}
D.~Tran, W.~Sheng, L.~Liu, and M.~Liu, ``{A Hidden Markov Model based driver
  intention prediction system},'' in \emph{IEEE International Conference on
  Cyber Technology in Automation, Control, and Intelligent Systems (CYBER)},
  2015, pp. 115--120.

\bibitem{Gindele2010}
T.~Gindele, S.~Brechtel, and R.~Dillmann, ``{A probabilistic model for
  estimating driver behaviors and vehicle trajectories in traffic
  environments},'' in \emph{13th International IEEE Conference on Intelligent
  Transportation Systems}, 2010, pp. 1625--1631.

\bibitem{Kahn2017}
G.~Kahn, A.~Villaflor, V.~Pong, P.~Abbeel, and S.~Levine, ``{Uncertainty-Aware
  Reinforcement Learning for Collision Avoidance},'' \emph{arXiv}, 2017.

\bibitem{Fan2019}
D.~D. Fan, J.~Nguyen, R.~Thakker, N.~Alatur, A.-a. Agha-mohammadi, and E.~A.
  Theodorou, ``{Bayesian Learning-Based Adaptive Control for Safety Critical
  Systems},'' in \emph{IEEE International Conference on Robotics and Automation
  (ICRA)}, 2020, pp. 4093--4099.

\bibitem{Blundell2015}
C.~Blundell, J.~Cornebise, K.~Kavukcuoglu, and D.~Wierstra, ``Weight
  uncertainty in neural networks,'' in \emph{International Conference on
  Machine Learning}, vol.~37.\hskip 1em plus 0.5em minus 0.4em\relax PMLR, Jul
  2015, pp. 1613--1622.

\bibitem{Lakshminarayanan2016}
B.~Lakshminarayanan, A.~Pritzel, and C.~Blundell, ``Simple and scalable
  predictive uncertainty estimation using deep ensembles,'' in \emph{Advances
  in Neural Information Processing Systems 30}, 2017, pp. 6402--6413.

\bibitem{Gal2016}
Y.~Gal and Z.~Ghahramani, ``Dropout as a bayesian approximation: Representing
  model uncertainty in deep learning,'' in \emph{International Conference on
  Machine Learning}, 2016.

\bibitem{Hu2018}
Y.~Hu, W.~Zhan, and M.~Tomizuka, ``{Probabilistic Prediction of Vehicle
  Semantic Intention and Motion},'' in \emph{IEEE Intelligent Vehicles
  Symposium}, vol. 2018-June, 2018, pp. 307--313.

\bibitem{Ding2019}
W.~Ding, J.~Chen, and S.~Shen, ``{Predicting vehicle behaviors over an extended
  horizon using behavior interaction network},'' in \emph{International
  Conference on Robotics and Automation (ICRA)}, 2019, pp. 8634--8640.

\bibitem{full_report}
R.~Cheng, R.~M. Murray, and J.~W. Burdick, ``{Limits of Probabilistic Safety
  Guarantees when Considering Human Uncertainty},'' \emph{arXiv}, 2021.

\bibitem{Pelckmans2008}
K.~Pelckmans, J.~D. Brabanter, J.~{A.K. Suykens}, and B.~{De Moor}, ``{Support
  and Quantile Tubes},'' \emph{arXiv}, 2008.

\bibitem{Ehrenfeucht1989}
A.~Ehrenfeucht, D.~Haussler, M.~Kearns, and L.~Valiant, ``{A general lower
  bound on the number of examples needed for learning},'' \emph{Information and
  Computation}, 1989.

\bibitem{arora2018}
S.~Arora, A.~Risteski, and Y.~Zhang, ``Do {GAN}s learn the distribution? some
  theory and empirics,'' in \emph{International Conference on Learning
  Representations}, 2018.

\bibitem{Bratman1992}
M.~E. Bratman, ``{Shared Cooperative Activity},'' \emph{The Philosophical
  Review}, 1992.

\bibitem{phan2019}
T.~Phan-Minh, K.~X. Cai, and R.~M. Murray, ``Towards assume-guarantee profiles
  for autonomous vehicles,'' in \emph{IEEE 58th Conference on Decision and
  Control (CDC)}, 2019, pp. 2788--2795.

\end{thebibliography}

\newpage

~

\newpage

%%% ------------------------------------------------------------- %%%
% \begin{appendix}

\section*{~~~~~~~~~~~~~~~~ Appendix A: \newline Fragility of Gaussian Uncertainty Model (Synthetic Data)}

We further tested the Gaussian uncertainty model on a synthetic 2D data set, using the same process detailed in Section 3A. Each 2D data point is analogous to a trajectory, $a=\tau_{[2:10]} \sim \mathcal{A}$, in the highD driving dataset. Therefore, the goal is still to learn the model $\hat{F}$ that best matches the data distribution $\mathcal{A}$, minimizing $m(\mathcal{A} || \mathcal{\hat{F}})$. However, using synthetic data allows us to test the accuracy of the uncertainty model with respect to a \textit{known} underlying probability distribution, $\mathcal{A}$.
%However, with synthetically generated data, we know the underlying distribution, $\mathcal{A}$, that all 2D points are drawn from. 

We randomly generated 10,000 2D points for training data (further increasing the amount of training data did not improve performance) from 3 different distributions: \textit{(a)} perfect Gaussian, \textit{(b)} Gaussian with uniform noise (magnitude of noise was $30\%$ of the data range), and \textit{(c)} Gaussian with symmetric non-uniform noise (also $30\%$ magnitude). For each of these training datasets, we computed the Gaussian uncertainty model that best fit the data. We then generated 10,000,000 2D points for our test data \textit{following the exact same distribution as the training data}, and observed how well our computed Gaussian uncertainty model captured the test data.

% \note{The orange line in Fig. \ref{fig:gaussian_toy} is perplexing.  While it is clear that the prediction accuracy is way off, one could interpret that extreme conservativeness of this model as in fact being successful--there are many fewer accidents than you might imagine.  You should probably discuss why the low number of errors is still a problem, if in fact it is.}   
\begin{figure}[h]
    \centering
    \includegraphics[width=0.46\textwidth]{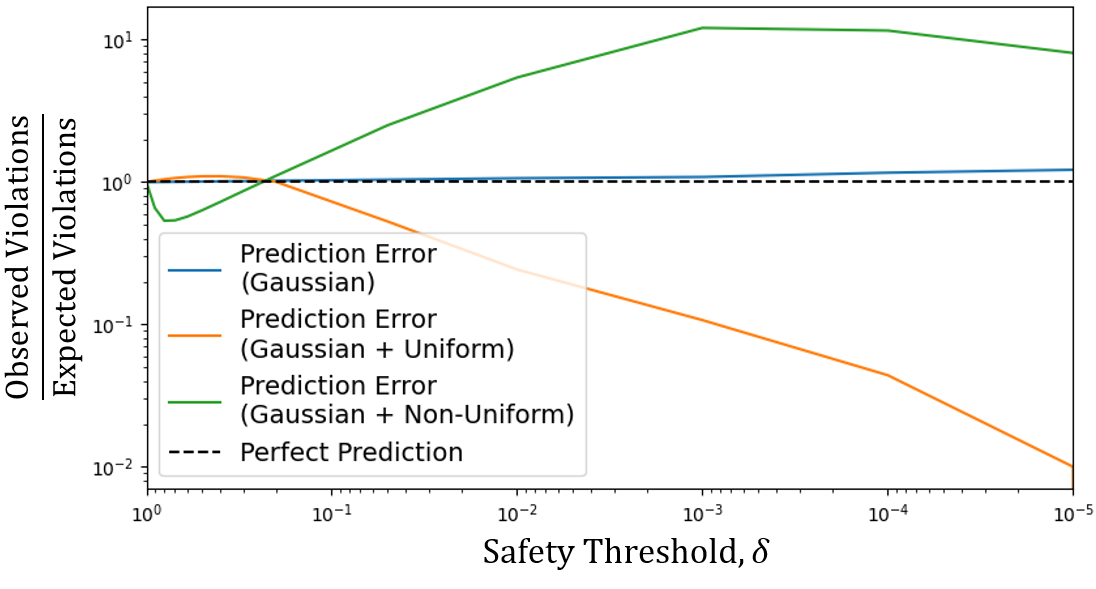}
    \caption{Prediction error vs. safety threshold, $\delta$, using a Gaussian uncertainty model on synthetic 2D data generated from 3 different distributions. The dashed black line represents a perfect prediction model. Significant prediction error arises when the underlying data distribution is non-Gaussian.}
    \label{fig:gaussian_toy}
\end{figure}

Figure \ref{fig:gaussian_toy} shows that the learned uncertainty model performed very well when the underlying data distribution was Gaussian (blue curve). However, it performed poorly (off by an order-of-magnitude) at low $\delta$ when the underlying distribution was non-Gaussian. When the underlying distribution was Gaussian with added uniform noise (orange curve), the observed violations were much lower than the expected violations (i.e. the model was conservative). This is good for safety, but would clearly lead to overly conservative behavior, especially since the model is off by orders of magnitude. 

However, more concerning is the case when the underlying distribution is Gaussian with \textit{non-uniform} noise (green curve). In this case, the observed violations were much higher than the expected violations (greater by an order of magnitude), posing a clear risk for safety-critical applications. This reinforces our results in Section 3A by illustrating that significant prediction error inevitably arises, regardless of the amount of training data, when the underlying data distribution is non-Gaussian.

\section*{~~~~~~~~~~~~~~~~ Appendix B: \newline Quantile Regression (Synthetic Data)}

We repeated the quantile regression experiments from Section 3C, using synthetic 2D data rather than real-world driving data. This allowed us to observe how well the uncertainty model performed under ideal conditions when the training/testing data were perfectly i.i.d. We randomly generated 1,000,000 2D training data points (analogous to 1,000,000 trajectories) following a Gaussian distribution, and computed $\delta$-quantile bounds following the same procedure described in Section 3C (i.e. computing the smallest convex set containing $1-\delta$ points). We then generated 10,000,000 2D test data points \textit{following the exact same distribution as the training data}, and observed how well our computed quantile bounds captured the test data.  
\begin{figure}[h]
    \centering
    \includegraphics[width=0.46\textwidth]{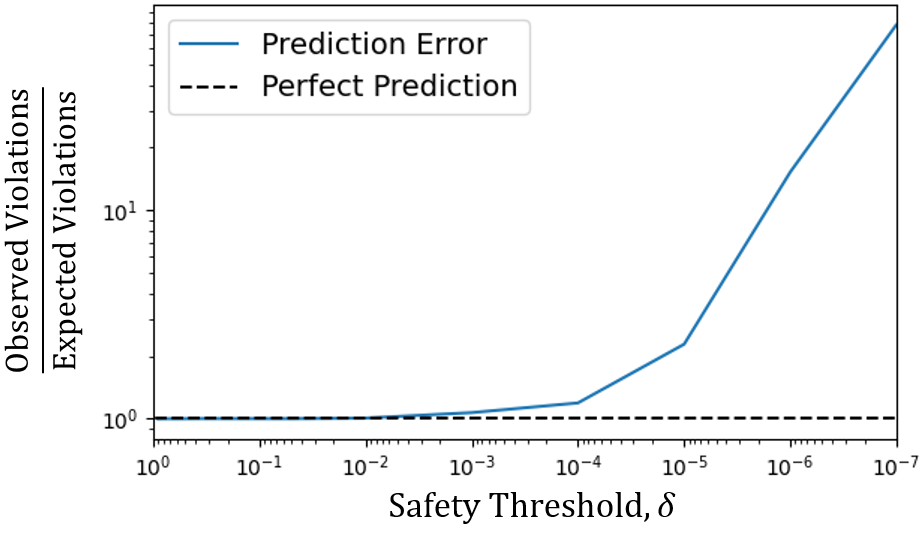}
    \caption{Prediction error vs. safety threshold, $\delta$ under computed quantile bounds on synthetic 2D data. The dashed black line represents a perfect prediction model.}
    \label{fig:quantile_toy}
\end{figure}

Figure \ref{fig:quantile_toy} shows the prediction error (i.e. ratio between expected and observed proportion of trajectories outside each quantile) versus the safety threshold $\delta$. The quantile regression model performed very well up to $\delta \geq 0.001$. However, performance rapidly deteriorated as $\delta$ decreased, meaning the model failed to accurately predict violation probabilities at those safety thresholds. This is consistent with our results in Section 3C.

% \note{Define when $\delta$ is "accurate" in the paragraph below}
% \note{The phrase "... when we project this inverse linear trend towards lower $\delta_{min}$ ..." is difficult to parse/read.  Can you rephrase?}

\begin{figure}
  \centering
  \subfigure[]{\label{fig:quantile_linear_toy}\includegraphics[width=0.42\textwidth]{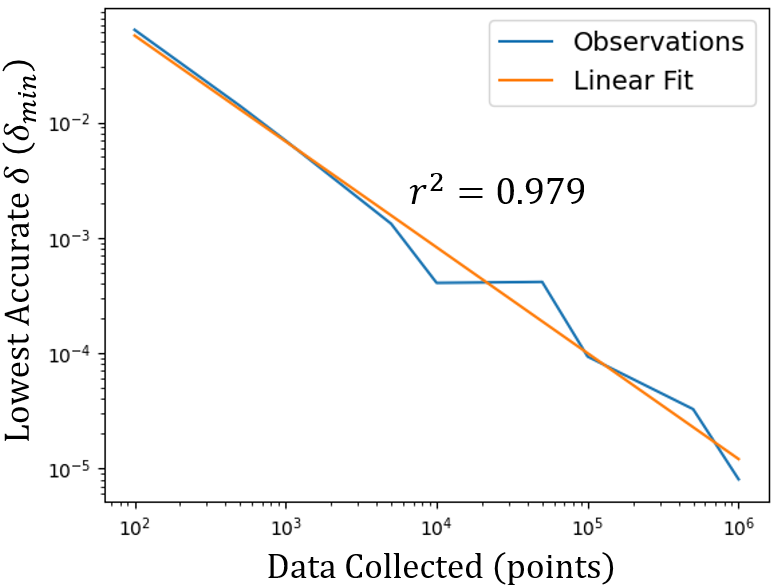}}\qquad
  \subfigure[]{\label{fig:quantile_scaling_toy}\includegraphics[width=0.42\textwidth]{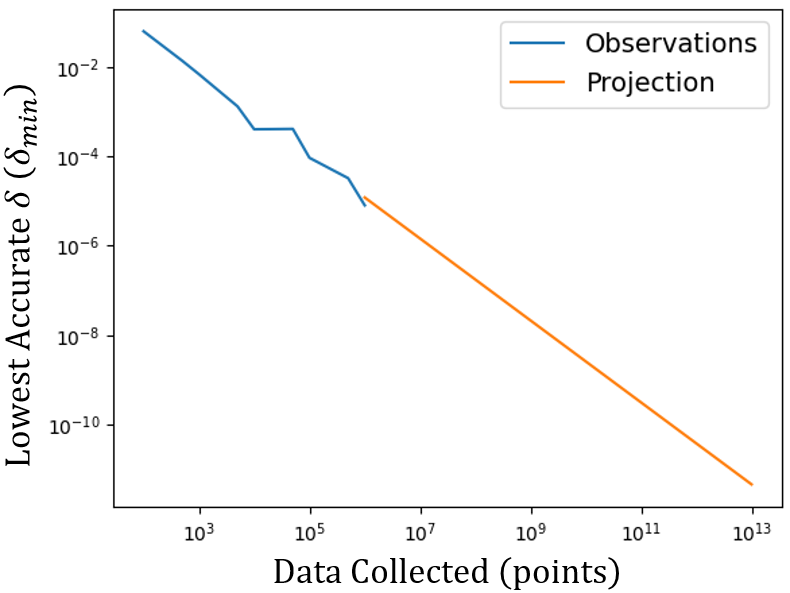}}
\caption{\textit{(a)} Smallest accurate $\delta$ versus amount of data using synthetic 2D data. The trend is highly linear ($r^2 = 0.979)$, \textit{(b)} Projection showing the expected amount of data required to obtain an accurate safety threshold $\delta_{min}$.}
\label{fig:quantile_scaling_appendix}
\end{figure}

Using the synthetic data, we computed the smallest accurate safety threshold, $\delta_{min}$, as a function of the amount of training data, $N$. This threshold $\delta_{min}$ was defined as follows,
\begin{equation}
    \delta_{min} ~ = ~ \min ~ \delta ~~~~~ \text{such that} ~~~~~  \left\lvert \log \Bigg( \frac{\text{expected}(\delta)}{\text{observed}(\delta)} \Bigg) \right\rvert \leq \varepsilon ~ .
\end{equation}
where we set $\varepsilon = 0.5$, which represents the vertical distance between the blue curve in Fig \ref{fig:quantile_toy} and the dotted black line. Therefore, $\delta_{min}$ represents the smallest $\delta$ such that our computed quantile bounds are $\varepsilon$-accurate (as described in Section 3C). Figure \ref{fig:quantile_linear_toy} shows the same inverse linear trend ($\delta_{min} \propto \frac{1}{N}$) on the synthetic data that was seen with the real driving data. Figure \ref{fig:quantile_scaling_toy} shows the extrapolation of this trend towards lower $\delta_{min}$. This result reinforces the point made in Section 3C that quantile regression can be very accurate for larger $\delta$, but it may not be feasible to collect enough data to reach safety thresholds $\delta_{min} \leq 10^{-8}$. 

\section*{~~~~~~~~~~~~~~~~ Appendix C: \newline Other Uncertainty Models}

\subsection{Generative Models}

Generative models have garnered significant interest in trajectory prediction for their ability to \textit{implicitly} learn the distribution $\mathcal{A}(x) = p(a | x)$. However, there are two significant issues with these approaches, the first of which is the time required to utilize these models in safety-critical situations. For example at best, a single prediction takes $\approx 0.05 s$ with Social-GAN \cite{Gupta2018}. In order to guarantee safety with probability $\delta = 0.01$, we would need to generate at least 100 trajectories taking $>5 s$. To guarantee safety with probability $\delta = 10^{-8}$, we would need to generate at least $10^{8}$ trajectories taking $>5,000,000 s$ ($> 1$ month), which is not suitable for real-time operation. While computational cost will surely decrease over time, it is unclear whether this modeling approach will be feasible in the near future.

More importantly, there are no guarantees that the uncertainty distribution implicitly captured by generative models provides any reasonable approximation to the true uncertainty distribution. It has been shown -- both empirically and theoretically -- that GANs can fail to learn the true distribution (suffering from ``mode collapse''), even when their training objective nears optimality \cite{arora2018}. Furthermore, the theoretical data efficiency bound described by Eq. \eqref{eq:scaling} suggests that the implicit distribution learned by such models will be inaccurate (at the safety thresholds we are considering) without currently infeasible amounts of data.  %In fact, the results shown in Figure \ref{fig:quantile} suggest that especially in the low $\delta$ regimes, these data-driven models perform poorly.

\subsection{Scenario Optimization Model}

Scenario Optimization is an appealing approach because (like quantile regression) it does not assume an underlying distribution over the data \cite{Sartipizadeh2020}. It relies only on the assumption that the data is drawn i.i.d. from some fixed (unknown) distribution. Therefore, we can obtain a high-confidence bound on the probability that a new trajectory is inside or outside a computed tube, \textit{without strong assumptions on the underlying distribution}. 

With this approach, the safety threshold, $\delta$, is a direct function of the amount of observed data \cite{Campi2018}; in other words $\delta = \delta(N)$, where $N$ is the number of training trajectories or ``samples''. Therefore, we cannot set arbitrarily small confidence levels (e.g. $\delta = 10^{-8}$). While this prevents users from applying the approach inappropriately, it requires very large amounts of data to get to low enough confidence levels for safety-critical applications. For example with $40,000$ trajectories, we were able to reach $\delta \approx 10^{-4}$ (after this point, computational feasibility became an issue). This suggests it is not feasible to reach desired $\delta$ levels given realistic datasets.

\begin{figure}
  \centering
  \subfigure[]{\includegraphics[width=0.45\textwidth]{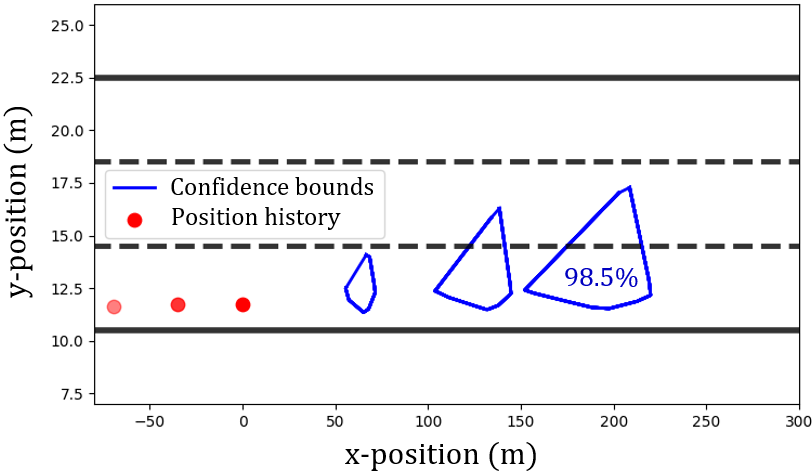}}\qquad
  \subfigure[]{\includegraphics[width=0.45\textwidth]{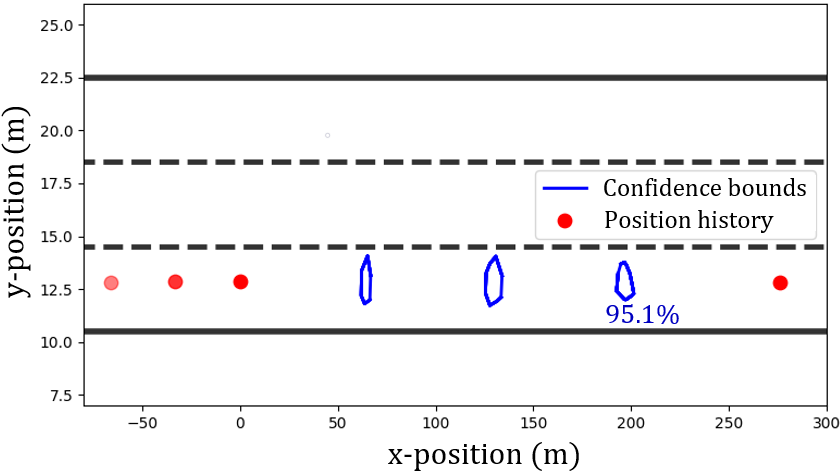}}
\caption{Plot of confidence bounds over the car's future trajectory. The car's positional history is shown by the red circles, and training data is taken from equivalent scenarios in the highD dataset. \textit{(a)} The goal position of the car is \textit{not} known. We compute a $98.5\%$ probability that a new trajectory falls within the blue confidence bounds at 2, 4, and 6 seconds in the future. \textit{(b)} The target position of the car \textit{is} known. We compute a $95.1\%$ probability that a new trajectory falls within the blue confidence bounds at 2, 4, and 6 seconds in the future.}
\label{fig:scenario_opt}
\end{figure}

Using the highD dataset and treating the trajectories in the training set as observed samples, we obtained high-confidence bounds (computed as the convex hull of the training trajectories) such that new trajectories should lie within those bounds with probability at least $1 - \delta$. For example, Figure \ref{fig:scenario_opt} shows the predicted confidence bounds for two representative driving instances; in Fig. \ref{fig:scenario_opt}a, the goal position is \textit{not} given but in in Fig. \ref{fig:scenario_opt}b, the goal position \textit{is} given. The scenario optimization approach predicts that the future trajectory of each car should fall within the blue confidence bounds at $2, 4, 6$ seconds in the future with $98.5\%$ (Fig. \ref{fig:scenario_opt}a) or $95.1\%$ (Fig. \ref{fig:scenario_opt}b) probability.

To test the accuracy of the computed confidence bounds, we examined how often trajectories in the test set actually remained within those bounds in the highD dataset. The ratio of observed violations to expected violations was smaller than expected (i.e. the method was conservative), which is reassuring for safety. Specifically, the observed vs. expected percentage of violations was approximately $5\%$ vs. $14\%$. 

However, the safety threshold $\delta(N)$ was always large ($\delta \in [0.02, 0.41]$) and unable to be arbitrarily defined, which makes the scenario optimization approach currently inapplicable to many safety-critical applications. This is consistent with our conclusion in Section 3C, that much more data is necessary to obtain reliable, probabilistic bounds.

\subsection{Hidden Markov Models}

Rather than reasoning about uncertainty only over trajectories, many methods in the POMDP literature reason about uncertainty over discrete intentions. Most often, these discrete intentions denote different goal positions for the agent, but they could also denote different operational modes (e.g. yield vs. no yield). Hidden Markov models enable us to compute an agent's most likely intention, which proves useful in solving many challenging problems. However, when guaranteeing safety with safety threshold $\delta$, intention must be correctly inferred with probability $1-\delta$. Issues arise when the intention must be inferred with very high confidence $\delta \leq 10^{-8}$.

\begin{figure}
  \centering
  \subfigure[Data for each mode generated from a Gaussian distribution.]{\includegraphics[width=0.4\textwidth]{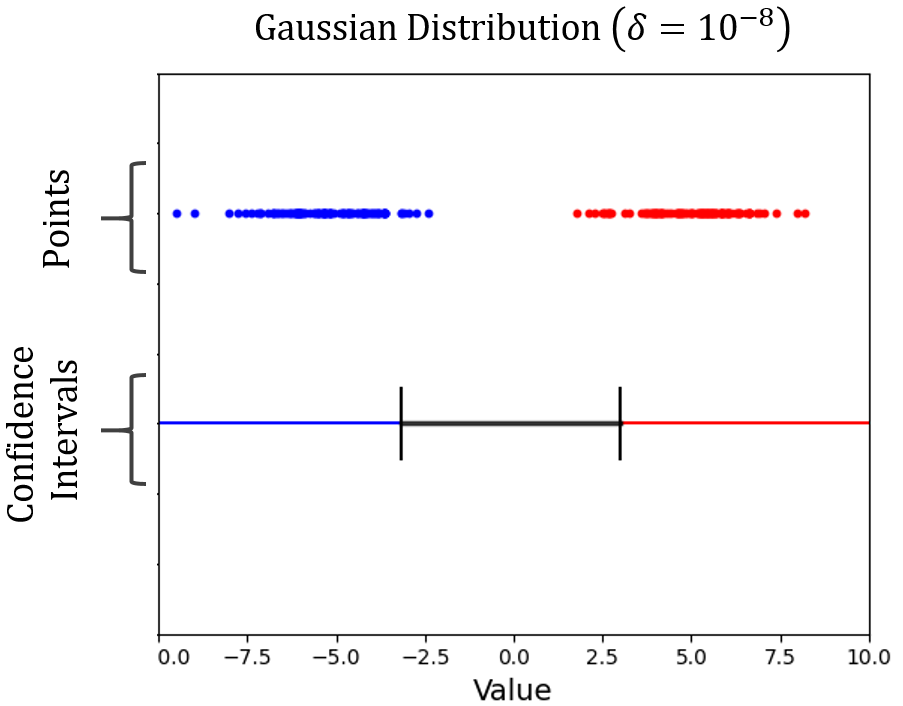}}\qquad
  \subfigure[Data for each mode generated from a uniform distribution.]{\includegraphics[width=0.4\textwidth]{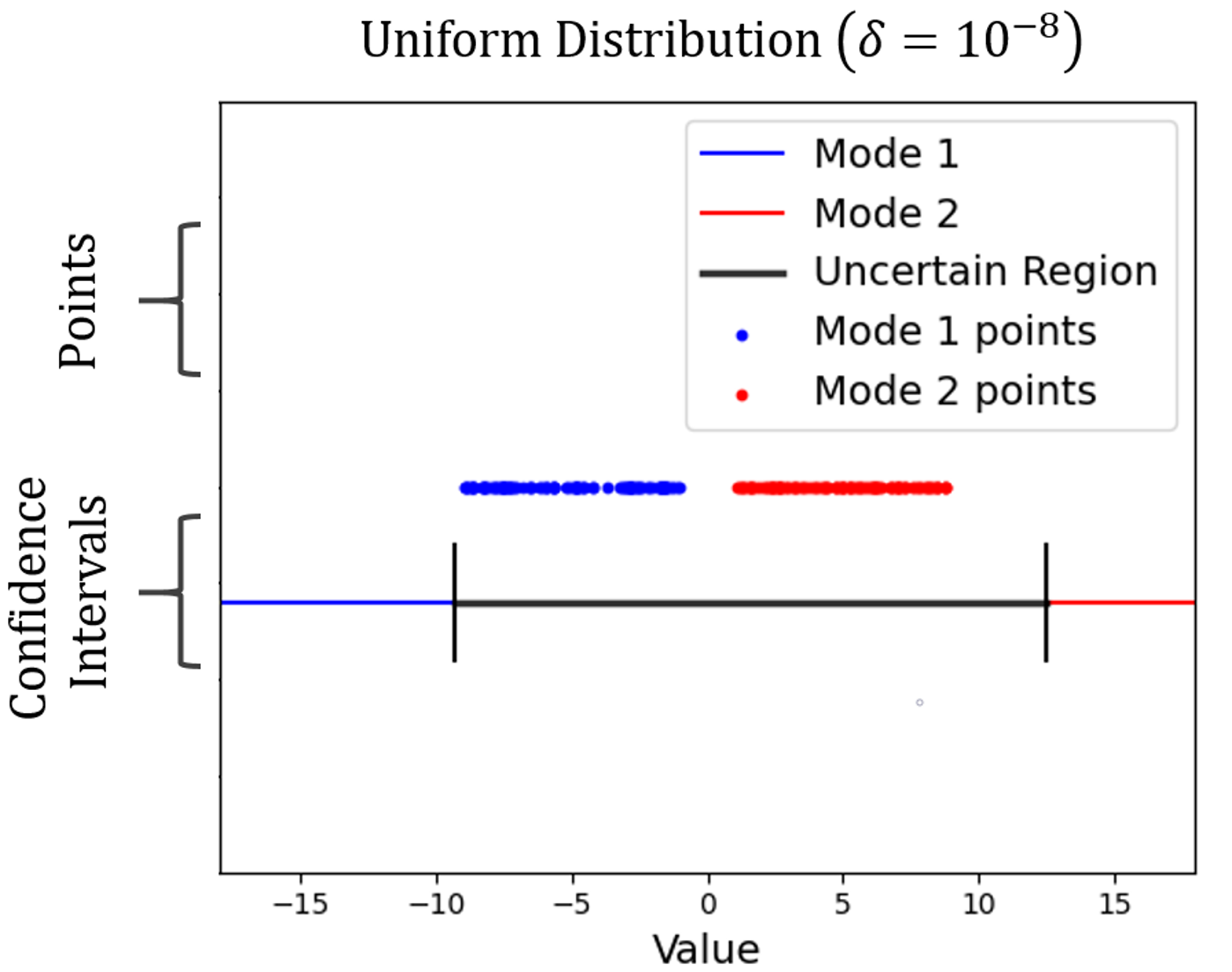}}
\caption{Synthetic data is generated from two different modes: \textit{(mode 1 -- blue, mode 2 -- red)}. The confidence intervals below denote where a point would have to lie in order to classify it, with confidence $\delta=10^{-8}$, as coming from either mode 1 or mode 2. For example, if a new point falls in the interval covered by the blue bar, it can be classified as coming from mode 1 with confidence $\delta\leq10^{-8}$. If it falls anywhere in the gray interval, we cannot conclude its mode (assuming a uniform prior).}
\label{fig:gaussian_intention}
\end{figure}

We demonstrate this on a 1D toy problem with synthetic data. We generated 1000 i.i.d. data points from two distinct distributions (mode 1 and mode 2), and computed the best fit Gaussian for each of these distributions. Note that our results did not change when increasing the amount of data. We then computed the intervals in which a new point would have to lie in order for us to classify it in either mode 1 or mode 2 with $1-\delta$ confidence. This was done by applying Bayes rule, assuming a uniform prior over the modes,
\begin{equation}
    \mathbb{P}(\text{mode} ~ | ~ x) = \frac{\mathbb{P}(x ~ | ~ \text{mode}) ~ \mathbb{P}(\text{mode})}{\mathbb{P}(x)} .
\end{equation}

Figure \ref{fig:gaussian_intention} shows these intervals when the points were generated from either a Gaussian distribution, or a uniform distribution. The interval covered by the gray line denotes the interval in which we can \textit{not} classify (with $\delta$ confidence) a point's mode. We note that the gray line extends across a significant portion of the data range, but is reasonable when the underlying distribution of points in each mode is \textit{perfectly Gaussian}. However, when the generated data is uniformly random, the uncertainty interval stretches across the entire range of data. This suggests that inferring intention or hidden ``modes'' under uncertainty will often be infeasible when considering very low safety thresholds, $\delta$, especially since we have shown that human behavioral variation is non-Gaussian. Furthermore, we cannot compensate for this non-Gaussian variation as we do not have accurate knowledge of the true distribution.

\end{document}